\documentclass[10pt,twocolumn,letterpaper]{article}

\usepackage{cvpr}              
\usepackage{algorithm}
\usepackage{algpseudocode}
\usepackage{pifont} 
\usepackage[symbol]{footmisc}
\definecolor{cvprblue}{rgb}{0.21,0.49,0.74}
\usepackage[pagebackref,breaklinks,colorlinks,allcolors=cvprblue]{hyperref}

\def\paperID{38028} 
\def\confName{CVPR}
\def\confYear{2026}

\title{Geometry-as-context: Modulating Explicit 3D in Scene-consistent \\ Video Generation to Geometry Context}

\author{
JiaKui Hu$^{1,2}$\footnotemark[1], Jialun Liu$^{2}$\textsuperscript{\textdagger, \textdaggerdbl}, Liying Yang$^{3,2}$, \\
Xinliang Zhang$^{1}$, Kaiwen Li$^{1}$, Shuang Zeng$^{1}$, Yuanwei Li$^{1}$, Haibin Huang$^{2}$, Chi Zhang$^{2}$, Yanye Lu$^{1}$\footnotemark[2] \\
$^1$Institute of Medical Technology, Peking University 
$^2$TeleAI $^3$MUST
}

\begin{document}

\twocolumn[{%
\maketitle
\vspace{-2em}
\centering
\includegraphics[width=0.95\linewidth]{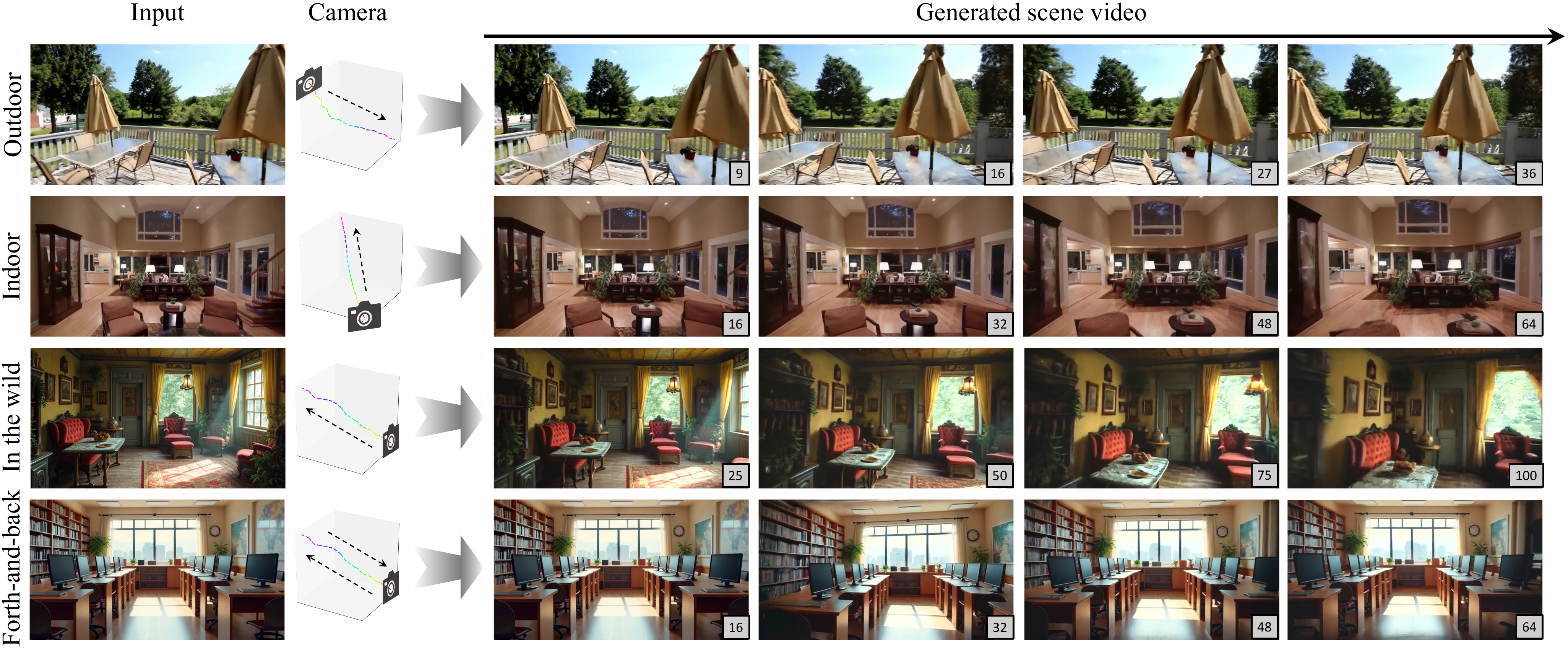}
\vspace{-3mm}
\captionof{figure}{Teaser demonstration. We introduce Geometry-as-Context (GaC), a framework that leverages explicit 3D information into reconstruction-based scene video generation. GaC mitigates cumulative errors from non-differentiable reconstruction and non-end-to-end training pipelines. Furthermore, GaC enhances the 3D consistency and long-term 3D memory of generative video models. We showcase GaC under four settings:on outdoor, indoor, in-the-wild, and forth-and-back camera trajectory. GaC maintains consistency under cyclic motion: even when an object (e.g., a computer) disappears in the 32-nd frame of the last row, it is faithfully restored in later frames.}\label{fig:teaser}
\vspace{1em}
}]

\footnotetext[1]{This work was done when he interned in TeleAI.}
\footnotetext[2]{Corresponding author.}
\footnotetext[3]{Project leader.}

\begin{abstract}
Scene-consistent video generation aims to create videos that explore 3D scenes based on a camera trajectory. Previous methods rely on video generation models with external memory for consistency, or iterative 3D reconstruction and inpainting, which accumulate errors during inference due to incorrect intermediary outputs, non-differentiable processes, and separate models. To overcome these limitations, we introduce ``geometry-as-context". It iteratively completes the following steps using an autoregressive camera-controlled video generation model: (1) estimates the geometry of the current view necessary for 3D reconstruction, and (2) simulates and restores novel view images rendered by the 3D scene. Under this multi-task framework, we develop the camera gated attention module to enhance the model's capability to effectively leverage camera poses. During the training phase, text contexts are utilized to ascertain whether geometric or RGB images should be generated. To ensure that the model can generate RGB-only outputs during inference, the geometry context is randomly dropped from the interleaved text-image-geometry training sequence. The method has been tested on scene video generation with one-direction and forth-and-back trajectories. The results show its superiority over previous approaches in maintaining scene consistency and camera control.
\end{abstract}

\section{Introduction}
\label{sec:intro}

Scene-consistent video generation aims to explore a scene with high 3D consistency starting from reference images, using a user-given camera trajectory, and returning its spatiotemporal RGB information. This task is significant for applications requiring 3D experiences or simulations, such as games, AR/VR, and embodied intelligence. The challenge of this task lies in ensuring that after any view transformation by the user, the geometry and texture of the same object in the simulated scene remain consistent.

Existing methods can be categorized into two primary classifications. Video-based methods~\cite{yu2023long,wang2024motionctrl,yu2025context,li2025vmem,xiao2025worldmem,he2025matrix} focus on achieving scene consistency solely through video generation models. Memory retrieval~\cite{xiao2025worldmem,li2025vmem,yu2025context} can preliminarily achieve this objective. However, it is still difficult to maintain 3D consistency in complex scenes and with large camera movements. 

In contrast, reconstruction-based methods~\cite{wiles2020synsin,rockwell2021pixelsynth,fridman2023scenescape,muller2024multidiff,yu2024viewcrafter,ni2025recondreamer,yu2025wonderworld} use explicit 3D signals to iteratively synthesize images from novel viewpoints. This process starts by predicting the geometry of the current view. Subsequently, this predicted geometry is used to reconstruct the 3D representation of the target scene, exemplified by structures such as the point cloud~\cite{yu2024viewcrafter} and the 3DGS~\cite{ni2025recondreamer}. The constructed 3D representation is utilized to render images within the specified camera poses. Due to the presence of texture degradations, such as voids or occlusions in the rendered images, an inpainting step is needed to produce photorealistic results. The inpainted images are then reintegrated into the above steps, allowing autoregressive updates to the existing 3D scene and inpainting of images for future views.

Reconstruction-based methods exhibit the capability to maintain robust scene consistency over limited spatial extents. However, their applicability to extended long ranges is hindered by unavoidable cumulative errors~\cite{li2025vmem}. These errors primarily originate from inaccuracies in reconstructions or inpaintings, compounded by the iterative processing. For long-range video generation affected by similar accumulated errors, it is feasible to mitigate these cumulative errors by training~\cite{chen2024diffusion,huang2025self}. Nevertheless, in reconstruction-based scene generation, the presence of \textit{(1) non-differentiable operators} in inverse rendering, coupled with \textit{(2) non-end-to-end training} between reconstruction and inpainting networks, makes it impractical to alleviate such errors through improved forcing methods~\cite{chen2024diffusion,huang2025self}.

We propose the \textit{geometry-as-context} (GaC) framework to mitigate the cumulative errors that arise in reconstruction-based scene video generation. The core idea is to replace the non-differentiable part in reconstruction-based scene video generation pipelines with a fully differentiable generative model. Leveraging the strong priors of generative models, GaC can estimate geometry, simulate novel-view rendering, and perform inpainting or restoration in a unified manner. To enable end-to-end optimization, we interleave the intermediate outputs—such as geometry and rendered images—into a single video sequence. An autoregressive camera-controlled video generation model is then used to synthesize this sequence, effectively unfolding and jointly training the iterative processes of geometry estimation, 3D reconstruction, rendering, and image inpainting.

However, directly implementing this idea presents two key challenges. First, existing camera-conditioning mechanisms (e.g., simple addition or concatenation) lack the capacity to simultaneously support multiple tasks such as geometry estimation and view synthesis. Second, interleaving RGB and geometry frames can confuse the model about which modality to generate next, potentially resulting in redundant or unstable geometric predictions. To address these issues, we introduce both a tailored architecture and a specialized training strategy for GaC. Architecturally, we propose a camera-gated attention mechanism that enhances the model’s camera controlability: camera poses are encoded in Plücker rays and used to enhance the query and produce a gating matrix that modulates self-attention outputs. This design helps the model distinguish how camera information should drive geometry prediction versus novel-view image synthesis. During training, a geometry dropout strategy is implemented, which randomly drops geometry contexts during the training process. This approach ensures that the model can learn the scene-consistency from the geometry context modeling, while also allowing it to bypass redundant geometry output during inference.

We evaluate the model on the task of scene video generation. Given a single reference image and a camera trajectory, the model generates high-fidelity scene videos while maintaining 3D consistency. Both quantitative and qualitative results demonstrate that the proposed method surpasses previous methods in terms of camera pose control and fidelity, and demonstrates robust generalization to scenes with large camera dynamics. Furthermore, ablation studies corroborate the effectiveness of the proposed framework, architecture, and training strategy.
\section{Related Work}

\textbf{Novel View Synthesis (NVS).} The advent of neural representations such as Neural Radiance Fields (NeRF)~\cite{mildenhall2021nerf} and 3D Gaussian Splatting (3DGS)~\cite{kerbl20233d} has significantly transformed the NVS task. \cite{fridovich2023k,song2023nerfplayer} primarily aims to reconstruct NeRF from synchronized multi-view images, a capability that remains inaccessible to the average user. \cite{gao2024gaussianflow,wang2025shape} leverage the efficiency of 3DGS to generate novel views from monocular video inputs, often integrating supplementary regularizations such as optical flow or depth information to enhance output quality. However, these methods are constrained to reconstructing observable regions, thereby producing artifacts when viewed from notably divergent views. Advanced generation methods~\cite{rombach2022high,sun2024autoregressive} are increasingly used in NVS due to their sufficient prior knowledge. The 3D object generation models~\cite{shi2023mvdream,liu2023zero,long2024wonder3d,gao2024cat3d,hu2025auto} use diffusion~\cite{rombach2022high} or autoregressive~\cite{sun2024autoregressive} models to generate novel views of the 3D object. For scene-level NVS, existing methods often incorporate conditions such as camera embeddings~\cite{wang2024motionctrl,li2025cameras}, depth-warped imaging~\cite{muller2024multidiff,ma2025you}, Plücker rays~\cite{he2024cameractrl,hu2025auto}, and point cloud renderings~\cite{yu2024viewcrafter} into the video diffusion model. 

\noindent \textbf{Scene-consistent Video Generation.}\label{sec:related_work} Recent studies leverage the capabilities of pre-trained video generation models~\cite{blattmann2023stable,yang2024cogvideox} to produce interactive, scene-level videos through the integration of camera or action control. They are dedicated to modulating the explicit 3D representation of scenes and refining the 3D scene consistency within the predicted frames of these models. The depth map, point cloud, and 3DGS serve as a source of explicit 3D information, which is instrumental in enhancing the consistency of generated scenes. SceneScape~\cite{fridman2023scenescape} employs an additional depth estimator to construct a mesh from a given image. Subsequently, it generates the projected novel views through diffusion models and then reprojects the novel views to refine and extend the mesh. Similarly, ViewCrafter~\cite{yu2024viewcrafter} executes an iterative method. However, it replaces depth maps with point clouds as explicit 3D. It utilizes a pre-trained reconstruction model to generate a point cloud and employs a video diffusion model to facilitate image inpainting or restoration4~\cite{hu2025universal,yao2026bridging}. GEN3C~\cite{ren2025gen3c} reconstructs point clouds with pixel-level depth maps, using them as the 3D cache for more precise camera manipulation during viewpoint extrapolation. Both Voyager~\cite{huang2025voyager} and TrajectoryCrafter~\cite{yu2025trajectorycrafter} leverage point clouds to maintain explicit 3D representation. Likewise, RenconDreamer~\cite{ni2025recondreamer} and Wonderworld~\cite{yu2025wonderworld} make use of 3DGS as explicit 3D representation. They also adopt an iterative approach, rendering and restoring novel view images using existing geometric structures, and subsequently employing these images to complete the 3DGS. To achieve more flexible and interactive scene generation, \cite{yu2025context} use the auto-regressive diffusion model to predict more scene-consistent images by selecting more relevant preceding views.


\section{Preliminary}

\subsection{Reconstruction-based Scene Video Generation}\label{sec:preliminaries}

Given an image $I_0\in \mathbb{R}^{H \times W \times 3}$, the scene video generation models expand the view based on the image $I_0$ and the camera poses $P_{i}, i\in [1, T]$ of $T$ novel views. The generated novel views need to maintain high 3D scene consistency while ensuring photorealistic quality.

Reconstruction-based methods primarily enforce the output images to satisfy 3D scene consistency by introducing explicit 3D representations. Based on related works in Sec.~\ref{sec:related_work} and above definitions, we can formulate reconstruction-based scene video generation methods as Algorithm~\ref{algo:scene_gen}. Through iterative reconstruction and next-view generation, the scene generation model can perform view expansion while ensuring 3D scene consistency. They use depth or normal maps, point clouds, 3DGS, etc., as 3D representations. The pre-trained large reconstruction model~\cite{wang2024dust3r,wang2025moge,wang2025vggt} is used as the geometry estimator $\epsilon(\cdot)$. The above steps can be represented by the Algorithm~\ref{algo:scene_gen} and the following formulas.

\begin{algorithm}[t]
\caption{\small Reconstruction-based Scene Video Generation}\label{algo:scene_gen}
\begin{algorithmic}[1]
\Require Reference image $I_0$
\Ensure $T$ novel view images.
\State Initialize view index $i \gets 0$
\While{i $\le$ T}
\State Estimate geometry information $G_i$ from image $I_i$ using geometry estimator $\epsilon(\cdot)$;
\State Unproject $G_i$ into a 3D representation, or update the existing 3D representation;
\State Render the 3D representation from pose $P_{i+1}$ into a novel view $I'_{i+1}$;
\State Inpaint or restore $I'_{i+1}$ into a photorealistic image $I_{i+1}$ using the generation model $\varrho(\cdot)$;
\State $i \gets i + 1$
\EndWhile
\end{algorithmic}
\end{algorithm}

\begin{align}
G_i &= \epsilon(I_i); \label{eq:geo} \\
\text{3D} &= \text{Unproject}(I_i, G_i); \label{eq:3d} \\
I'_{i+1} &= \text{Render}(\text{3D}, P_{i+1}); \label{eq:render} \\
I_{i+1} &= \varrho(I'_{i+1} , P_{i+1}). \label{eq:inpaint}
\end{align}

\noindent \textbf{Discussion: cumulative errors.} Although these methods can generate high-consistent scene images, they are also affected by cumulative errors. When the geometric estimator $\epsilon(\cdot)$ fails to provide accurate geometric information, the resulting 3D representation will deviate from the real scene, further impacting subsequent rendering and novel view generation. This error will gradually expand with the increase in iterations, much like the ``butterfly effect", ultimately leading to the ambiguous scene data.

This cumulative error is more severe than the cumulative error in long-range video generation. The latter can be alleviated through an autoregressive rolling~\cite{huang2025self}, while the former is difficult to improve through learning due to its \textbf{\textit{non-differentiability}}. This is because we use two independent models for geometry prediction and novel view inpainting, respectively. The non-differentiable unprojection and rendering operations in the algorithm make the gradients cannot be transmitted between the two models. This also makes end-to-end training of these models impossible.

\section{Method}

\begin{figure*}[ht]
\centering
\includegraphics[width=\linewidth]{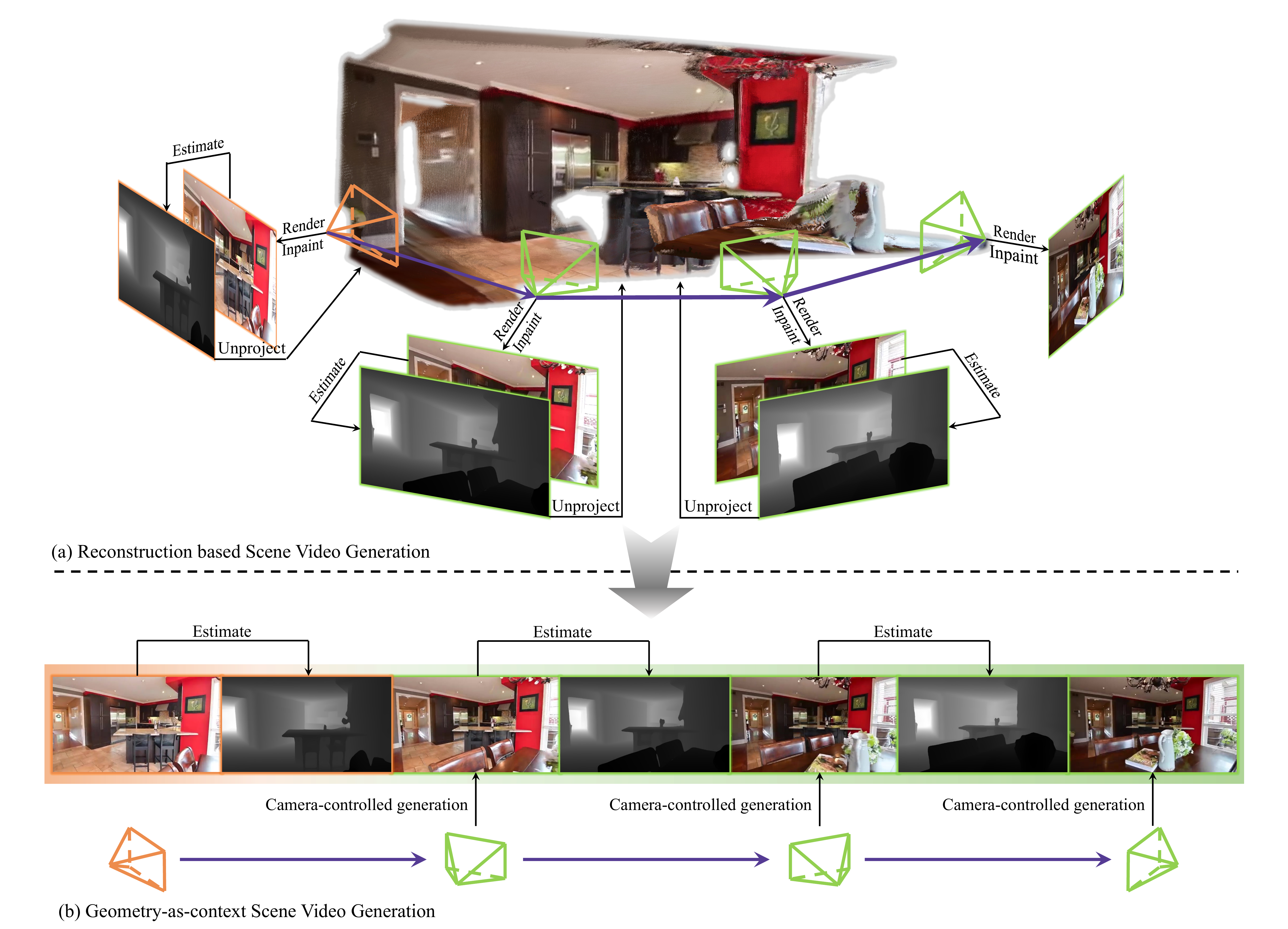}
\caption{Reconstruction-based scene video generation (a) v.s. our geometry-as-context (GaC) (b). Reconstruction-based scene video generation uses non-differentiable operators in reconstruction, which tend to worsen cumulative errors caused by inaccurate geometry estimates or image inpainting. In contrast, GaC replaces these operations with camera-controllable generation, turning reconstruction-based scene video generation into an autoregressive video generation framework with one single DiT. It can effectively reduce cumulative errors caused by non-differentiable reconstruction and non-end-to-end training.} 
\label{fig:pipeline}
\end{figure*}

To address these cumulative errors, we propose to internalize the geometry estimation, reconstruction, and rendering operators into one model. We aim to unify these subtasks in Algorithm~\ref{algo:scene_gen}, thereby alleviating the cumulative errors caused by separate models and non-differentiable operators.

\subsection{Geometry-as-context}\label{sec:unify_recon}

We integrate non-differentiable in Eq.~\ref{eq:3d} within the generation model framework. Using geometric information, \textit{e.g.} depth maps, as contexts, the iterative algorithm in Algorithm~\ref{algo:scene_gen} can be flattened into an autoregressive model.

Specifically, we merge Eq.~\ref{eq:3d} with Eq.~\ref{eq:render} as follows.

\begin{equation}
I'_{i+1} = \text{Func}(I_i, G_i, P_{i+1}), 
\end{equation}

\noindent where $\text{Func}(\cdot)$ is a function composed of Eq.~\ref{eq:3d} and Eq.~\ref{eq:render}. It satisfies $\text{Func}(I_i, G_i, P_{i+1}) = \text{Render}(\text{Unproject}(I_i, G_i), P_{i+1})$.

In this scenario, the function $F(\cdot)$ can be represented by a video generation model with camera and geometry control. We use a differentiable generation model $\varphi(\cdot)$ to replace the non-differentiable Eq.~\ref{eq:3d}. The task of the model $\varphi(\cdot)$ is to warp the previous image $I_i$ by its geometric information $G_i$ and the camera pose $P_{i+1}$ of the target view, similar to the depth-warped imaging in~\cite{fridman2023scenescape,yu2024viewcrafter,huang2025voyager}. This adjustment can help alleviate the cumulative error exacerbated by non-differentiable operators. In summary, the equations for scene generation can be reformulated as follows.

\begin{align}
G_i &= \epsilon(I_i); \label{eq:geo_pred_2} \\
I'_{i+1} &= \varphi(I_i, G_i, P_{i+1}), \label{eq:image_warp_2} \\
I_{i+1} &= \varrho(I'_{i+1} , P_{i+1}). \label{eq:image_comp_2}
\end{align}

In the above equations, we use three independent networks for geometric prediction (Eq.~\ref{eq:geo_pred_2}), image warping (Eq.~\ref{eq:image_warp_2}), and image completion (Eq.~\ref{eq:image_comp_2}), respectively. However, the model parameters used are separate, which may still exacerbate cumulative errors. We explore a basic solution: using a \textbf{single model} to handle the above three steps in a unified manner. In other words, we reformulate the above equations in the following form.

\begin{equation}\label{eq:unify}
\{G_i, I'_{i+1}, I_{i+1}\} = \varrho(\{I_i, G_i, I'_{i+1}\}, P_{i+1}),
\end{equation}

\noindent where $\varrho$ is a camera controlled generation model. 

The model $\varrho$ can process multiple images \textbf{in sequence} and output the required results for the corresponding tasks. Following ReCamMaster~\cite{bai2025recammaster}, we adopt the frame-dimension concatenation for parallel processing images $\{I_i, G_i, I'_{i+1}\}$ due to its ease of implementation and better performance. When the images $\{I_i, G_i, I'_{i+1}\}$ are concatenated along the frame dimension, Eq.~\ref{eq:unify} admits an autoregressive manner with a context length of 1. This formulation can be further generalized to accommodate longer contexts, as shown in Figure~\ref{fig:pipeline} (b).

Note that there is a clear sequence among these subtasks. For example, when performing image warping (Eq.~\ref{eq:image_warp_2}), we rely on the predicted geometric information as a condition. This means that before predicting the warped image, we need to first predict the geometric information of the given image. Therefore, the sequence of frame inputs that are fed into the model $\varrho$ should be strictly controlled as follows: the $i$-th frame, the geometry of the $i$-th frame, the $i$-th warped image, the $(i+1)$-th frame, etc.

\noindent \textbf{Variants.} Eq.~\ref{eq:unify} can be further combined to produce the following three variants.

\textit{Variant \#1: geometry as context.} By combining image warping (Eq.~\ref{eq:image_warp_2}) and image completion (Eq.~\ref{eq:image_comp_2}), Eq.~\ref{eq:unify} can be reformulated as follows. 

\begin{equation}\label{eq:gac}
\{G_i, I_{i+1}\} = \varrho(\{I_i, G_i\}, P_{i+1}).
\end{equation}

In Eq.~\ref{eq:gac}, the model $\varrho$ estimates geometry information $G_i$, subsequently generating the RGB image $I_{i+1}$ for the next view. Firstly, compared to Eq.~\ref{eq:unify}, this variant results in a reduced sequence length, thus improving efficiency during the training process. Secondly, the inclusion of explicit geometric estimation operations enables the model to acquire the ability to recognize three-dimensional information, consequently augmenting the 3D scene consistency of the generated images. Lastly, the two targets that require generation exist in distinct modalities and exhibit significant differences in texture and form, which allows the model to distinctly identify the subsequent tasks to be addressed with minimal prompting. Consequently, we employ this variant as the training scheme.

\textit{Variant \#2: warped image as context.} By combining geometry prediction (Eq.~\ref{eq:geo_pred_2}) and image warping (Eq.~\ref{eq:image_warp_2}), Eq.~\ref{eq:unify} can be reformulated as follows.

\begin{equation}\label{eq:wac}
\{I'_{i+1}, I_{i+1}\} = \varrho(\{I_i, I'_{i+1}\}, P_{i+1}).
\end{equation}

In Eq.~\ref{eq:wac}, the model $\varrho$ simulates point cloud rendered image $I'_{i+1}$ while generating image $I_i$ with clean textures. This variant eschews the comprehension of 3D geometric information entirely, which consequently complicates the extrapolation of 3D scene-consistency. As a result, this variant is not employed in the training process. 

\textit{Variant \#3: without context.} In this extreme case, Eq.~\ref{eq:unify} can degenerate into the following:

\begin{equation}\label{eq:video_gen}
\{I_{i+1}\} = \varrho(I_i, P_{i+1}).
\end{equation}

Under this variant, the model is effectively reduced to a purely video generation framework, resulting in the complete omission of the 3D reconstruction component. Consequently, this leads to the forfeiture of the capability to explicitly achieve 3D consistency. Therefore, this variant is not used during the training process.

\subsection{Architecture}

Since we perform multiple tasks within a model, we need to enable the model to discern the distinct functions that camera poses fulfill in the contexts of geometry prediction and novel view synthesis. We design a more precise camera control to meet this need.

\begin{figure}[ht]
\centering
\includegraphics[width=\linewidth]{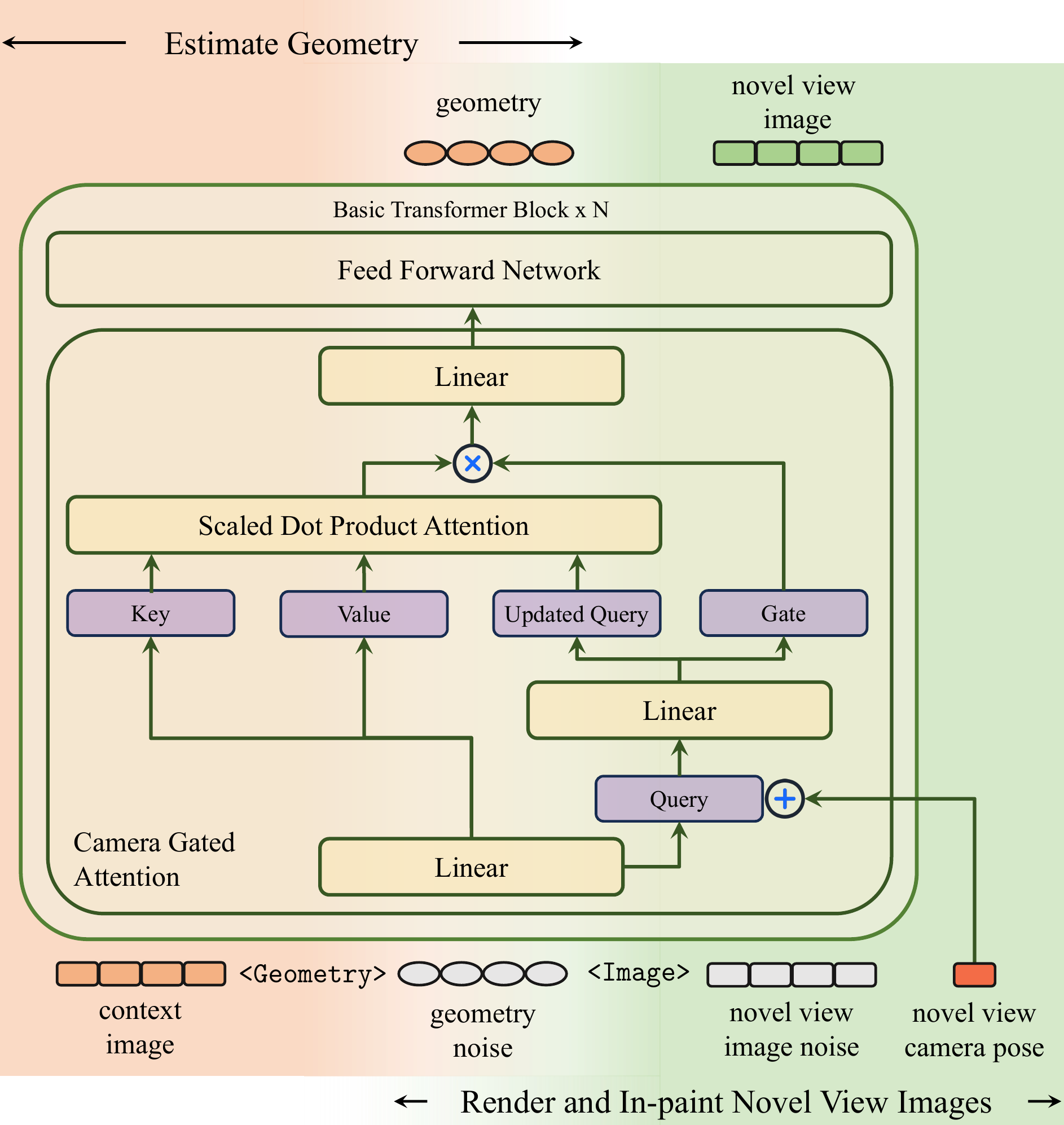}
\vspace{-3mm}
\caption{Detailed architecture of geometry-as-context.}\label{fig:arch}
\end{figure}

\noindent \textbf{Camera gated attention (CGA)} modulates the query of self-attention through the incorporation of camera pose features, subsequently generating gates to regulate the output of the self-attention layer. Given a latent feature $F_{i} \in \mathbb{R}^{N \times D}$ of the image $I_i$ and patchified Plücker rays $r_{i} \in \mathbb{R}^{N \times D}$ of the target camera pose $P_{i}$, CGA processes them as follows.

\begin{align}
\{Q, K, V\} &= \text{Linear}_1 (F_{i}); \\
\{Q_{res}, Gate\} &= \text{Linear}_2 (Q + r_{i}); \\
O &= \text{SDPA} (Q + Q_{res}, K, V); \\
O &= \text{Linear}_3 (O * \sigma(Gate)),
\end{align}

\noindent where $\sigma(\cdot)$ is the sigmoid function and $\text{SDPA}(\cdot)$ is the scaled dot product attention. We use a convolution to patch-ify the Plücker rays. 

The CGA integrates camera pose information into the self-attention mechanism, thereby enabling the model to incorporate pose information for both feature querying and output gating. This integration facilitates the model's ability to differentiate the distinct roles that camera pose information plays in various applications, including geometry estimation and novel view synthesis.


\subsection{Training}\label{sec:unify_depth}

Under the autoregressive framework in Figure~\ref{fig:pipeline} (b), we need to tell the model when to perform geometry and when to generate images. Otherwise, the model may be confused about the output format, leading to a decline in performance. 

\noindent \textbf{Text-guided interleaved modeling.} As shown in Figure~\ref{fig:arch}, we use text to inform the model about the next task it will process. Specifically, we design the input sequence in an interleaved form.

\begin{equation}\nonumber
\{ I_{i}, \texttt{<Geometry>}, G_{i}, \texttt{<Image>}, I_{i+1}, \cdots \}
\end{equation}

This sequence helps the model's acquisition of task-related information based on text contexts. Meanwhile, employing human language for task scheduling enables users to actively intervene in the generation process.

However, this interleaved text-image-geometry sequence presents new challenges related to the efficiency of training and inference. (1) The presence of geometry contexts results in reduced training efficiency by doubling the length of the sequence. (2) Inference outputs may be redundant if the user does not require geometry output. We address these challenges by implementing a strategy of randomly dropping geometry context during the training phase.

\begin{table*}[ht]
\centering
\resizebox{\linewidth}{!}{
\begin{tabular}{lcccccccccccccc}
\toprule
& \multicolumn{6}{c}{\textbf{RealEstate10K~\cite{zhou2018stereo}}} & \multicolumn{6}{c}{\textbf{Tanks-and-Temples~\cite{knapitsch2017tanks}}} \\
\cmidrule(lr){2-7} 
\cmidrule(lr){8-13} 
\multicolumn{1}{l}{Method} & PSNR $\uparrow$ & SSIM $\uparrow$ & LPIPS$\downarrow$ & FID$\downarrow$ & $R_{err}$$\downarrow$ & $T_{err}$$\downarrow$ & PSNR $\uparrow$ & SSIM $\uparrow$ & LPIPS$\downarrow$ & FID$\downarrow$ & $R_{err}$$\downarrow$ & $T_{err}$$\downarrow$ \\
\midrule
CameraCtrl~\cite{he2024cameractrl} & 15.93 & 0.550 & 0.440 & 77.93 & \textbf{0.020} & \textbf{0.266} & 12.28 & 0.389 & 0.642 & 201.72 & 0.080 & 0.533 \\
ViewCrafter~\cite{yu2024viewcrafter} & 16.72 & 0.585 & 0.417 & 80.47 & 0.022 & 0.327 & 12.59 & 0.438 & 0.549 & 115.16 & \textbf{0.058} & \textbf{0.399} \\
GEN3C~\cite{ren2025gen3c} & 18.12 & 0.624 & 0.402 & 66.20 & 0.027 & 0.344 & 15.32 & 0.506 & 0.509 & 90.52 & 0.074 & 0.430 \\
SEVA~\cite{zhou2025stable} & 17.42 & 0.601 & 0.431 & 73.95 & 0.021 & 0.284 & 15.60 & 0.518 & \textbf{0.490} & \textbf{88.76} & 0.072 & 0.477 \\
Voyager~\cite{huang2025voyager} & 18.70 & 0.616 & 0.395 & 65.12 & 0.035 & 0.596 & 15.24 & 0.487 & 0.510 & 90.06 & 0.081 & 0.515 \\
\textbf{GaC (Ours)} & \textbf{19.01} & \textbf{0.656} & \textbf{0.354} & \textbf{55.76} & 0.024 & 0.270 & \textbf{15.77} & \textbf{0.532} & 0.507 & 93.29 & 0.072 & 0.442 \\
\bottomrule
\end{tabular}
}
\caption{Quantitative results of scene video generation from single view with given camera trajectory.}
\label{table:single}
\vspace{-3mm}
\end{table*}

\noindent \textbf{Geometry Dropout.} Given an interleaved image-geometry sequence, we randomly remove the geometry context in the sequence based on a specified ratio $r \in (0, 1)$. When the geometry context is removed, the preceding and following frames satisfy the \textit{Variant \#3} (Eq.~\ref{eq:video_gen}). In this framework, the model directly predicts the novel view image based on the previous frame. This method effectively reduces the sequence length while maintaining the model's capacity to acquire explicit 3D consistency from geometry modeling. Therefore, it facilitates the model's ability to generalize across image-to-image transformations, thereby enabling purely image-contexted generation during the inference phase.

It is noteworthy that, despite employing the geometry dropout, the model is capable of producing the geometry of the present view via the \texttt{<Geometry>} prompt. This offers certain advantages for applications that require simultaneous video generation and 3D reconstruction. To substantiate this claim, we provide some test samples in the appendix.

Due to the varied geometry dropout rates in each sample, the sequence length across samples is inconsistent. We need to employ Fully Sharded Data Parallel (FSDP) as the model parallelization strategy and maintain a batch size of one.


\begin{figure*}[!ht]
\centering
\includegraphics[width=\linewidth]{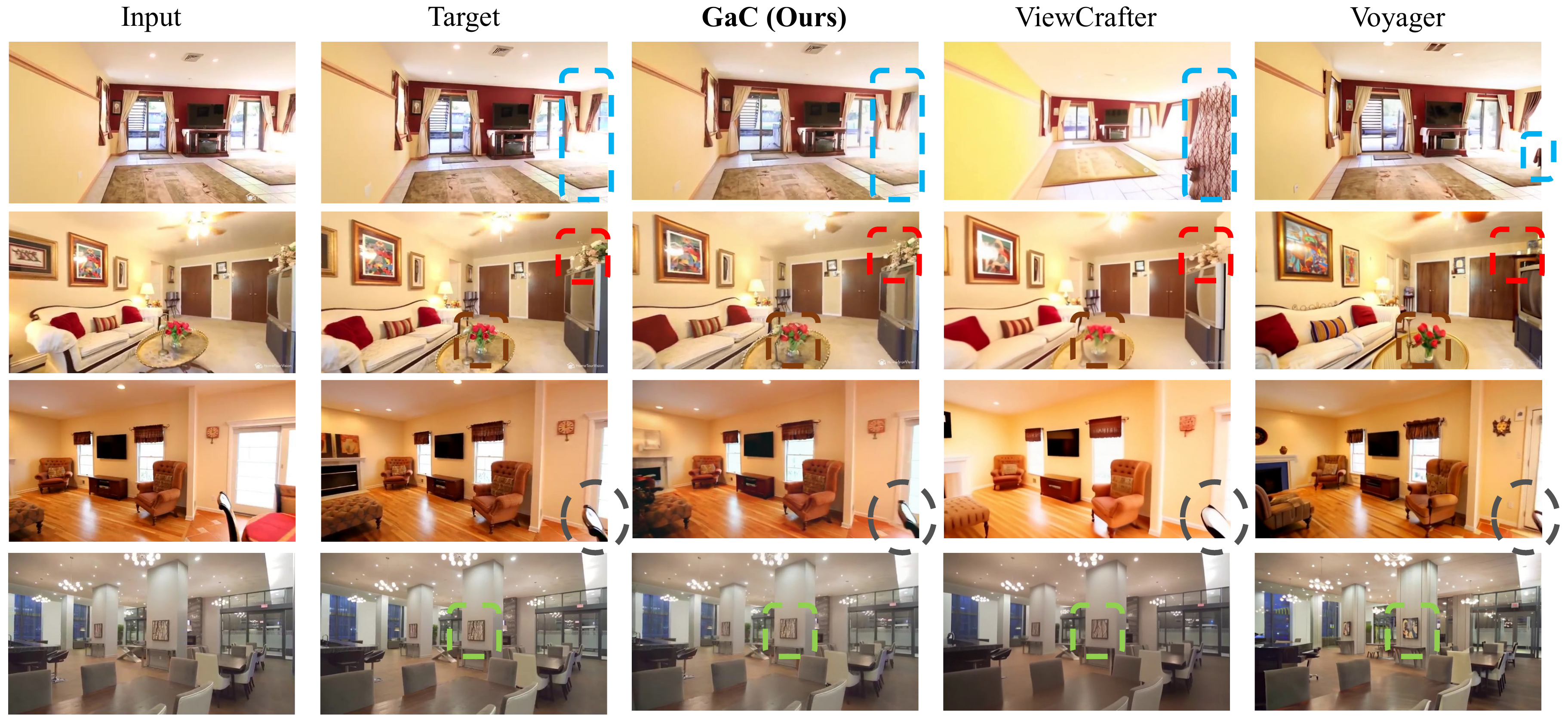}
\vspace{-5mm}
\caption{Qualitative results of scene video generation from single view. Compared to the baselines, our model generates more consistent novel views. These images are the 20-th frame of the generated video clip except the input one. For a clearer visualization, please zoom in.}\label{fig:single_view}
\vspace{-3mm}
\end{figure*}

\section{Experients}

\subsection{Experimental setup}

\noindent \textbf{Implementation details.} We use Bagel-7B~\cite{deng2025emerging} as the base model due to its text-image interleaved modeling capability. Following VMem~\cite{li2025vmem}, we train our model on 8-frame sequences. We randomly select the first 1-4 frames as context views and the subsequent frames as target views. We combine images captured from four consecutive views into a single frame in a grid pattern~\cite{hu2025omni}, which enhances the consistency between the various viewpoints. The resolution of each view is $640 \times 352$. The images are encoded as latent using FLUX-VAE~\cite{flux2024}. The model needs to be trained on 8 H100 for 40,000 iterations, which is expected to take two days. During inference, we select context views using the context-as-memory~\cite{yu2025context} approach. We do not use classifier-free guidance techniques during sampling.

\noindent \textbf{Datasets.} We train our model on 66033 video clips from RealEstate10K~\cite{zhou2018stereo} train set. We randomly select 150 video clips from RealEstate10K~\cite{zhou2018stereo} test set and Tanks-and-Temples~\cite{knapitsch2017tanks} for testing. The latter has larger camera pose motions. To collect the depth maps used during training, we improve the depth estimation strategy provided by Hunyuan-Voyager~\cite{huang2025voyager}, which is presented in the appendix.

\noindent \textbf{Evaluation.} During inference, the camera trajectory incorporates both the trajectory provided in the dataset and a forth-and-back trajectory. Specifically, in the forth-and-back trajectory, the initial 32 frames utilize the camera trajectory provided by the dataset, while the subsequent 32 frames return the camera to the camera pose of the first view. This forth-and-back trajectory serves to validate the model's scene-consistency. The evaluation metrics are: (1) Fréchet Image Distance (FID) for the assessment of video quality; (2) Peak Signal-to-Noise Ratio (PSNR), Structural Similarity Index Metric (SSIM), and Learned Perceptual Image Patch Similarity (LPIPS) for pixel-wise discrepancies between different views; (3) Alignment between camera poses of predicted views and their ground truth, as follows~\cite{he2024cameractrl}. We use VGGT~\cite{wang2025vggt} to extract poses from the predicted views. The camera accuracy is calculated using the formula from prior research~\cite{li2025vmem}.

\subsection{Main results}

\noindent \textbf{Scene video generation from single-view.} We first propose quantitative results in Table~\ref{table:single}. GaC demonstrates superior performance compared to other models in a range of metrics. Specifically, the reduced FID values suggest that our model is more accurately aligned with the intended data distribution. Furthermore, lower LPIPS scores, in conjunction with higher PSNR and SSIM values, indicate enhanced visual quality and improved pixel-level and structural fidelity relative to the targets. Additionally, our approach exhibits improved precision in camera control compared to video-based models, as evidenced by $R_{err}$ and $T_{err}$.

Qualitative comparisons are presented in Figure~\ref{fig:single_view}. Compared to the previous reconstruction-based scene video generation method~\cite{yu2024viewcrafter}, our approach excels in visual quality. This is evidenced by enhanced color fidelity, as observed in the color of the wall in the first row of Figure~\ref{fig:single_view}, and the presence of more intricate texture details, exemplified by the flowers' texture in the second row of Figure~\ref{fig:single_view}. Additionally, our approach offers improved precision in camera manipulation than the video generation-based method~\cite{huang2025voyager}, illustrated by the chair backrest in the third row of Figure~\ref{fig:single_view}, and ensures greater consistency in inter-frame textures, as seen in the texture of the picture located centrally in the last row of Figure~\ref{fig:single_view}.

\noindent \textbf{Scene video generation on the forth-and-back trajectory.} Quantitative results are presented in Table~\ref{table:cycle} (next page). The GaC continues to demonstrate superior numerical performance relative to other methods. However, it is observed that, in forth-and-back camera trajectories, the efficacy of all methods experiences a marked decline. Consequently, the development of advanced context memory strategies emerges as a crucial area for enhancing scene consistency. Due to the page limit, we put the qualitative comparisons in the appendix.

\begin{table*}[ht]
\centering
\resizebox{\linewidth}{!}{
\begin{tabular}{lcccccccccccccc}
\toprule
& \multicolumn{6}{c}{\textbf{RealEstate10K~\cite{zhou2018stereo}}} & \multicolumn{6}{c}{\textbf{Tanks-and-Temples~\cite{knapitsch2017tanks}}} \\
\cmidrule(lr){2-7} 
\cmidrule(lr){8-13} 
\multicolumn{1}{l}{Method} & PSNR $\uparrow$ & SSIM $\uparrow$ & LPIPS$\downarrow$ & FID$\downarrow$ & $R_{err}$$\downarrow$ & $T_{err}$$\downarrow$ & PSNR $\uparrow$ & SSIM $\uparrow$ & LPIPS$\downarrow$ & FID$\downarrow$ & $R_{err}$$\downarrow$ & $T_{err}$$\downarrow$ \\
\midrule
CameraCtrl~\cite{he2024cameractrl} & 13.80 & 0.391 & 0.501 & 98.32 & 0.044 & 0.325 & 11.42 & 0.282 & 0.769 & 171.88 & 0.110 & 0.0766 \\
ViewCrafter~\cite{yu2024viewcrafter} & 15.77 & 0.437 & 0.430 & 72.14 & \textbf{0.042} & \textbf{0.411} & 13.96 & 0.358 & 0.640 & 133.66 & 0.134 & \textbf{0.576} \\
GEN3C~\cite{ren2025gen3c} & 15.28 & 0.524 & 0.431 & 80.03 & 0.081 & 0.454 & 13.80 & 0.321 & 0.648 & 129.91 & 0.098 & 0.778 \\
SEVA~\cite{zhou2025stable} & 15.12 & 0.528 & 0.472 & 85.39 & 0.048 & 0.422 & 14.00 & 0.349 & 0.620 & 101.62 & 0.100 & 0.662 \\
Voyager~\cite{huang2025voyager} & 15.80 & 0.521 & 0.409 & 79.81 & 0.077 & 0.638 & 14.06 & 0.337 & \textbf{0.652} & \textbf{89.37} & 0.120 & 0.650 \\
\textbf{GaC (Ours)} & \textbf{16.34} & \textbf{0.547} & \textbf{0.399} & \textbf{64.31} & 0.050 & 0.429 & \textbf{14.29} & \textbf{0.359} & 0.629 & 101.24 & \textbf{0.093} & 0.638 \\
\bottomrule
\end{tabular}
}
\caption{Quantitative results of scene video generation from single view with forth-and-back camera trajectory.}
\label{table:cycle}
\vspace{-3mm}
\end{table*}


\subsection{Ablation studies}

We conduct ablation studies on the RealEstate10K test set to verify the effectiveness of our proposed method.

\noindent \textbf{Variants.} In Section~\ref{sec:unify_recon}, we outline three variants of geometry-as-context. We conduct ablation studies on these variants. The variation \#3 shows the least effective performance due to its exclusive reliance on the autoregressive video generation model to ensure the scene consistency. Variant \#2 incorporates warped images as contextual information, resulting in a slight performance improvement over Variant \#3. However, the lack of 3D geometric constraints during training suggests that there is room for further enhancement. Adding geometric context helps the model to better learn a 3D representation of the scene, leading to improved performance in scene video generation.

\begin{table}[ht]
\centering
\resizebox{\linewidth}{!}{
\begin{tabular}{ccccccc}
\toprule
Context & PSNR $\uparrow$ & SSIM $\uparrow$ & LPIPS$\downarrow$ & FID$\downarrow$ & $R_{err}$$\downarrow$ & $T_{err}$$\downarrow$ \\
\midrule
None (\textit{Varient \#3}) & 16.34 & 0.551 & 0.412 & 89.03 & 0.029 & 0.351 \\
Warped image (\textit{Varient \#2}) & 18.33 & 0.671 & 0.383 & 59.12 & 0.024 & 0.299 \\
Geometry (\textit{Varient \#1}) & 19.01 & 0.656 & 0.354 & 55.76 & 0.024 & 0.270 \\
\bottomrule
\end{tabular}
}
\caption{Ablation on different variants of geometry-as-context.}
\label{tab:ablation_context}
\vspace{-3mm}
\end{table}

\noindent \textbf{Choices of geometry.} Besides depth maps, point maps can also be unprojected into 3D representations. We compare the impact of different geometry representations on scene video generation performance. The results are shown in Table~\ref{tab:ablation_geometry}. In this experiment, the FLUX-VAE model is used for data processing. The variation in geometry context types has a minimal effect on performance. However, depth maps perform slightly better than point maps. This can be attributed to the smaller modality gap between depth maps and natural images, which allows the VAE trained on natural images to more effectively extract geometric features from depth maps. In contrast, point maps generally require more precise encoding processes~\cite{xu2025geometrycrafter}.

\begin{table}[ht]
\centering
\resizebox{0.8\linewidth}{!}{
\begin{tabular}{ccccccc}
\toprule
Choices & PSNR $\uparrow$ & SSIM $\uparrow$ & LPIPS$\downarrow$ & FID$\downarrow$ & $R_{err}$$\downarrow$ & $T_{err}$$\downarrow$ \\
\midrule
Point map & 19.04 & 0.648 & 0.381 & 56.37 & 0.027 & 0.303 \\
Depth map & 19.01 & 0.656 & 0.354 & 55.76 & 0.024 & 0.270 \\
\bottomrule
\end{tabular}
}
\caption{Ablation on different choices of geometry.}
\label{tab:ablation_geometry}
\vspace{-3mm}
\end{table}

\noindent \textbf{Effect of the Camera Gated Attention (CGA).} We verify the effectiveness of CGA. Specifically, the application of CGA results in a marked reduction in both rotation error ($R_{err}$) and translation error ($T_{err}$). This indicates that the model is capable of producing images with camera poses that exhibit a higher degree of congruence with the ground truth images.

\begin{table}[ht]
\centering
\resizebox{0.75\linewidth}{!}{
\begin{tabular}{ccccccc}
\toprule
CGA & PSNR $\uparrow$ & SSIM $\uparrow$ & LPIPS$\downarrow$ & FID$\downarrow$ & $R_{err}$$\downarrow$ & $T_{err}$$\downarrow$ \\
\midrule
\ding{56} & 18.57 & 0.581 & 0.461 & 68.42 & 0.032 & 0.469 \\
\ding{52} & 19.01 & 0.656 & 0.354 & 55.76 & 0.024 & 0.270 \\
\bottomrule
\end{tabular}
}
\caption{Effect of the CGA.}
\label{tab:ablation_cga}
\vspace{-3mm}
\end{table}

\noindent \textbf{Effect of the geometry dropout.} The introduction of the geometry dropout technique has demonstrably resulted in a substantial reduction in both the training and inference cost, with a negligible decline in performance. The evaluation of training and inference time is performed on the H100.

\begin{table}[ht]
\centering
\resizebox{\linewidth}{!}{
\begin{tabular}{ccccccccc}
\toprule
Geometry & PSNR $\uparrow$ & SSIM $\uparrow$ & LPIPS$\downarrow$ & FID$\downarrow$ & $R_{err}$$\downarrow$ & $T_{err}$$\downarrow$ & Training & Inference \\
Dropout & & & & & & & Time & Time \\
\midrule
\ding{56} & 19.23 & 0.660 & 0.342 & 57.18 & 0.025 & 0.248 & 24 s/step & 4.6 s/img \\
\ding{52} & 19.01 & 0.656 & 0.354 & 55.76 & 0.024 & 0.270 & 11 s/step & 2.2 s/img \\
\bottomrule
\end{tabular}
}
\caption{Effect of the geometry dropout. The unit ``s/step" means how long it takes to perform one iteration. Unit ``s/img" means how long it takes to generate one image during inference.}
\label{tab:ablation_geo_dropout}
\vspace{-3mm}
\end{table}
\section{Conclusion}

This study attributes the cumulative inference errors in reconstruction-based scene video generation methods to two primary causes: (1) the presence of non-differentiable 3D reconstruction and rendering operators, and (2) the lack of end-to-end training arising from the decoupling of geometric prediction and novel view generation modules. To address these issues, we propose the Geometry-as-Context (GaC) framework. By leveraging generative models for geometric estimation and simulating the novel view rendering process, GaC effectively eliminates non-differentiable operations during inference. Furthermore, by unfolding the iterative processes of geometry prediction, view rendering, and inpainting, GaC enables fully end-to-end training through autoregressive video generation over interleaved RGB–geometry sequences. Experimental results demonstrate that GaC produces videos with high visual fidelity and improved 3D consistency.


\clearpage
{
    \small
    \bibliographystyle{ieeenat_fullname}
    \bibliography{main}
}

\clearpage
\setcounter{page}{1}
\maketitlesupplementary

\section{More results}

\subsection{Indoor scenes} 
\begin{figure*}[!ht]
\centering
\begin{tabular}{cccc}
\includegraphics[width=0.24\textwidth]{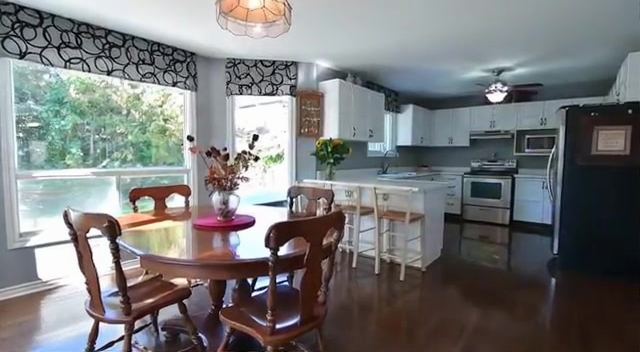} \\
Reference \\
\includegraphics[width=0.24\textwidth]{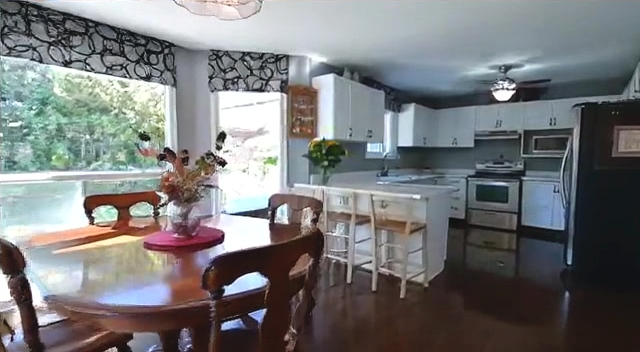} &
\includegraphics[width=0.24\textwidth]{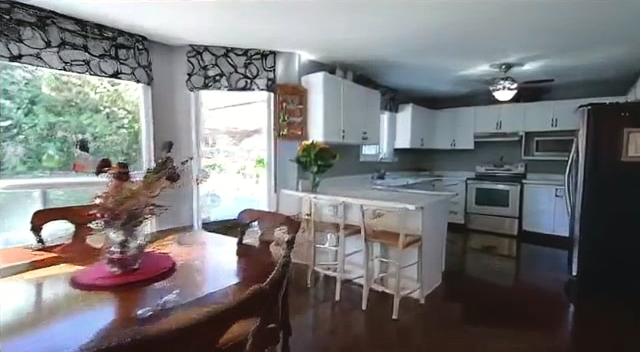} &
\includegraphics[width=0.24\textwidth]{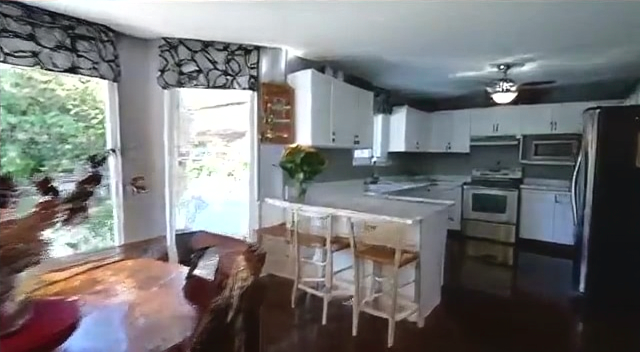} &
\includegraphics[width=0.24\textwidth]{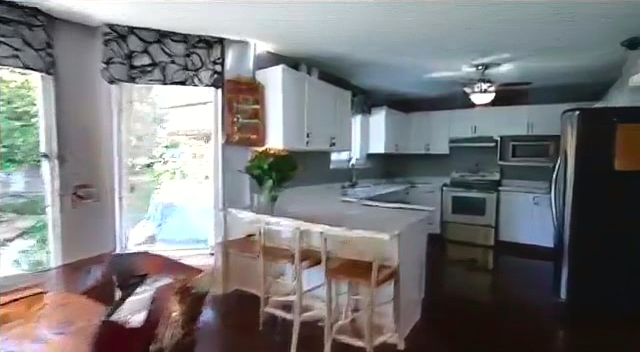} \\
View-16 & View-32 & View-48 & View-64 \\
\includegraphics[width=0.24\textwidth]{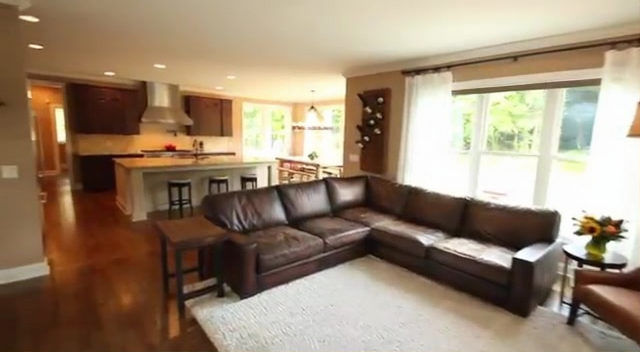} \\
Reference \\
\includegraphics[width=0.24\textwidth]{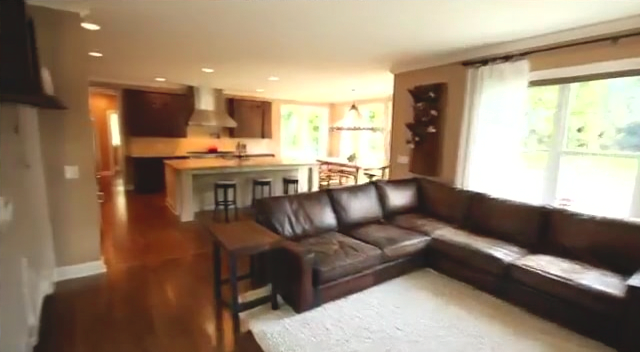} &
\includegraphics[width=0.24\textwidth]{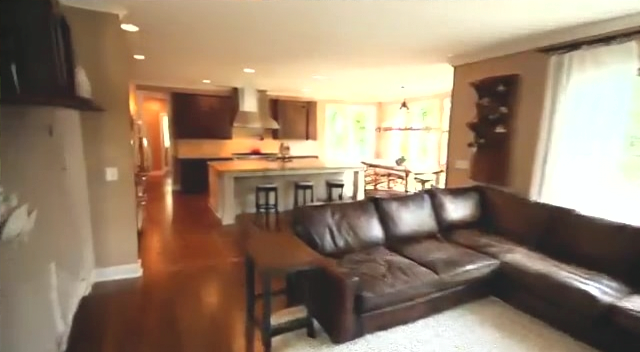} &
\includegraphics[width=0.24\textwidth]{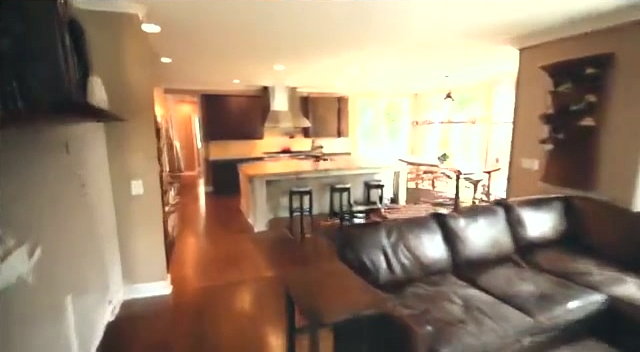} &
\includegraphics[width=0.24\textwidth]{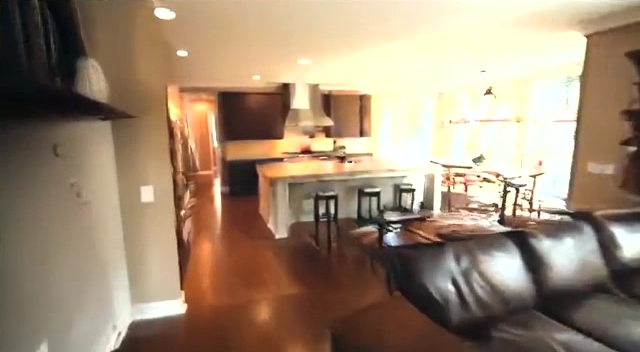} \\
View-16 & View-32 & View-48 & View-64 \\
\includegraphics[width=0.24\textwidth]{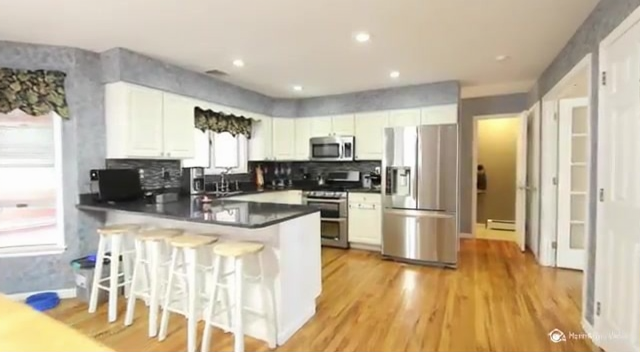} \\
Reference \\
\includegraphics[width=0.24\textwidth]{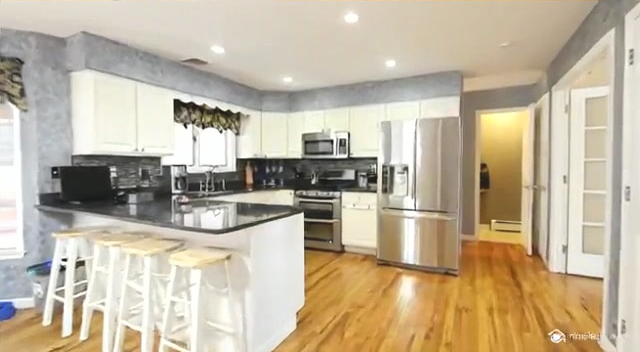} &
\includegraphics[width=0.24\textwidth]{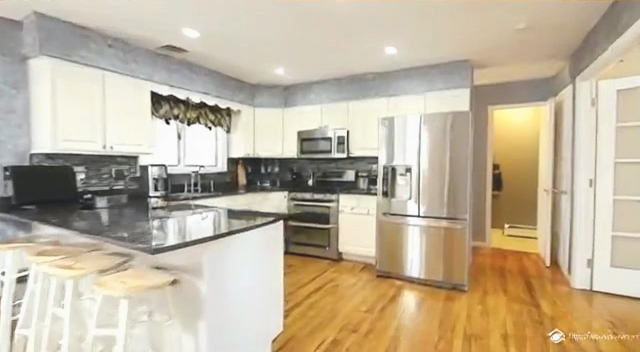} &
\includegraphics[width=0.24\textwidth]{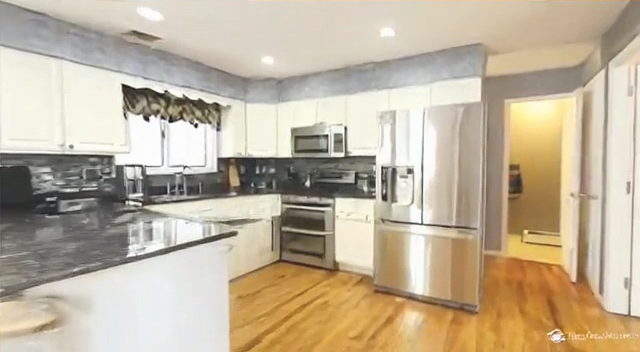} &
\includegraphics[width=0.24\textwidth]{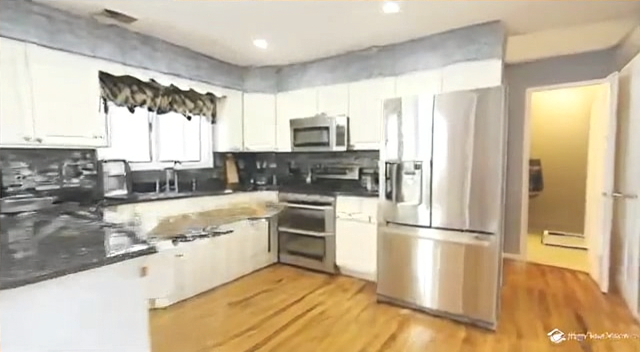} \\
View-16 & View-32 & View-48 & View-64 \\
\includegraphics[width=0.24\textwidth]{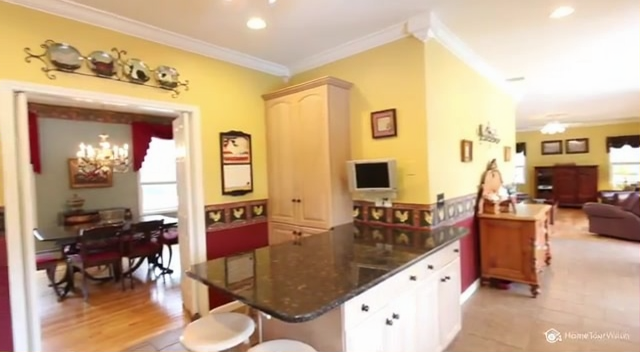} \\
Reference \\
\includegraphics[width=0.24\textwidth]{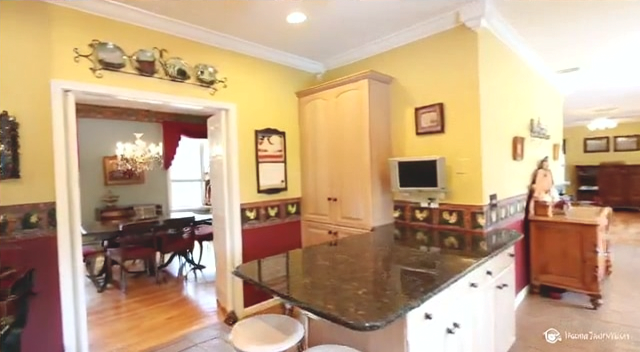} &
\includegraphics[width=0.24\textwidth]{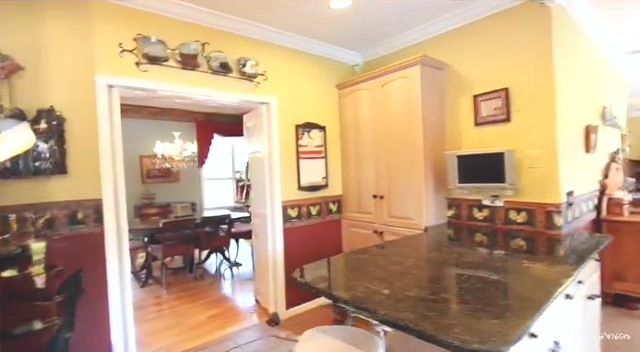} &
\includegraphics[width=0.24\textwidth]{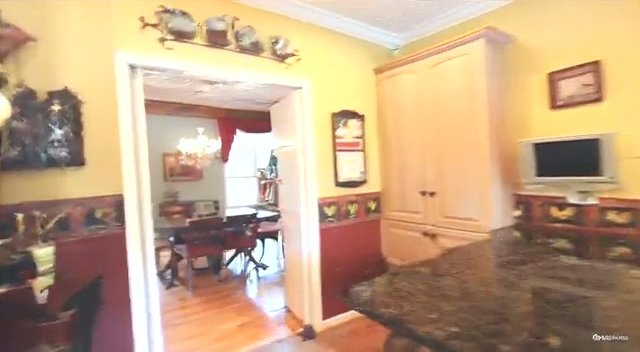} &
\includegraphics[width=0.24\textwidth]{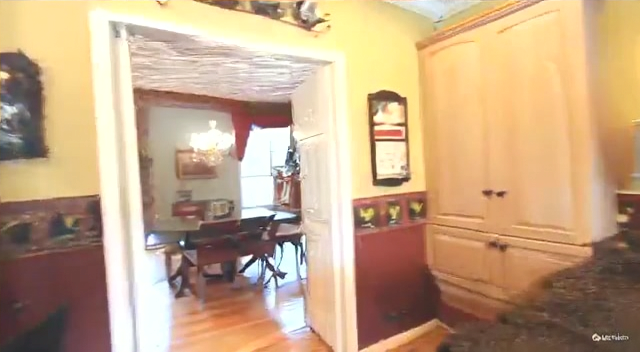} \\
View-16 & View-32 & View-48 & View-64 \\
\end{tabular}
\caption{Gac's results on indoor scenes.}\label{fig:indoor}
\end{figure*}

Please see Figure~\ref{fig:indoor}.

\subsection{Outdoor scenes} 
\begin{figure*}[!ht]
\centering
\begin{tabular}{cccc}
\includegraphics[width=0.24\textwidth]{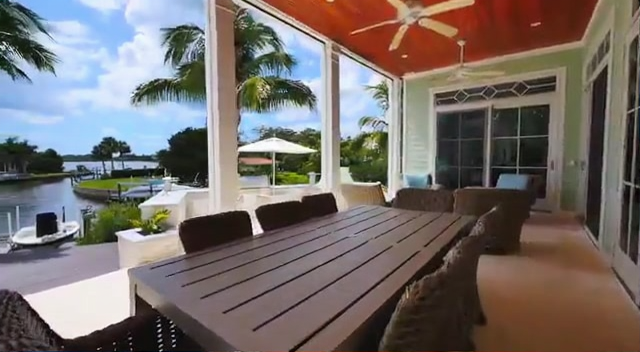} \\
Reference \\
\includegraphics[width=0.24\textwidth]{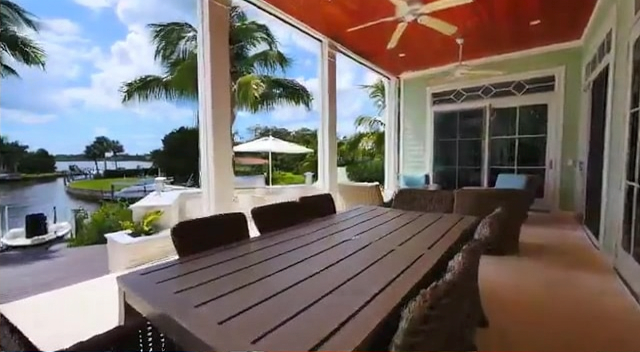} &
\includegraphics[width=0.24\textwidth]{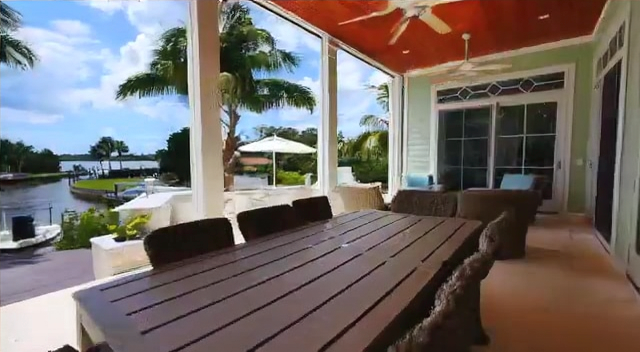} &
\includegraphics[width=0.24\textwidth]{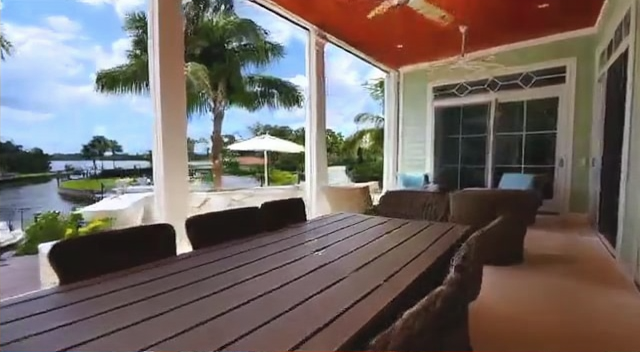} &
\includegraphics[width=0.24\textwidth]{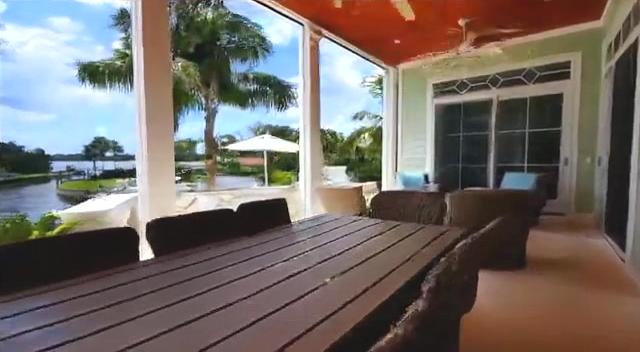} \\
View-6 & View-12 & View-18 & View-24 \\
\includegraphics[width=0.24\textwidth]{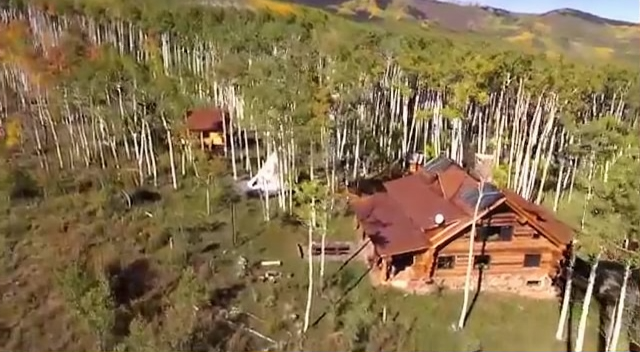} \\
Reference \\
\includegraphics[width=0.24\textwidth]{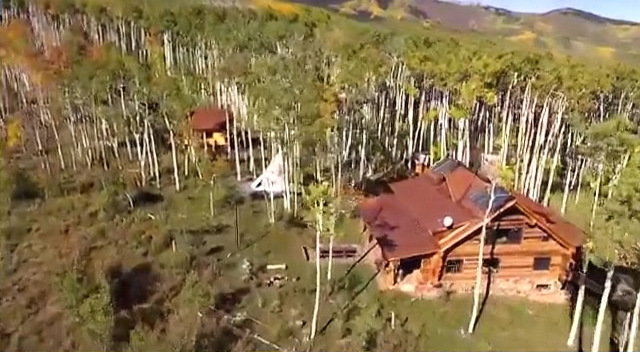} &
\includegraphics[width=0.24\textwidth]{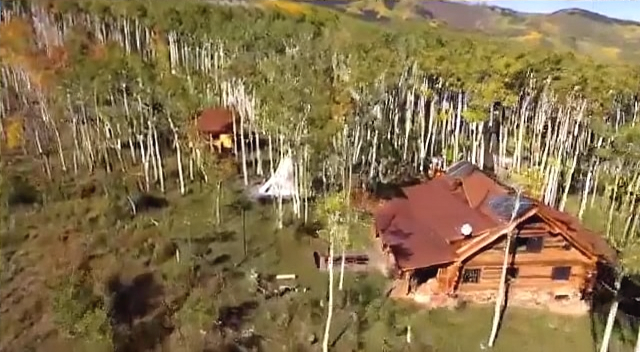} &
\includegraphics[width=0.24\textwidth]{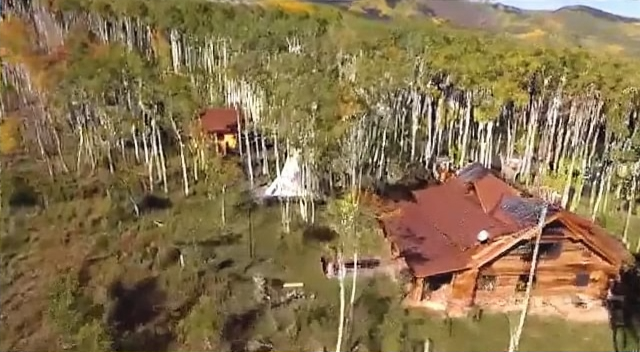} &
\includegraphics[width=0.24\textwidth]{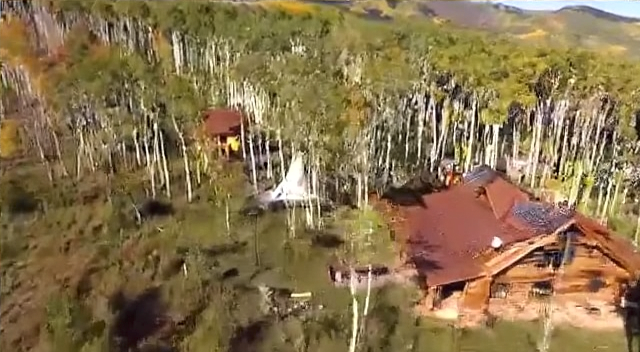} \\
View-6 & View-12 & View-18 & View-24 \\
\includegraphics[width=0.24\textwidth]{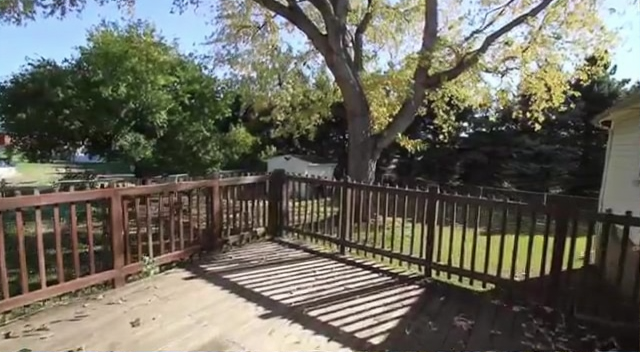} \\
Reference \\
\includegraphics[width=0.24\textwidth]{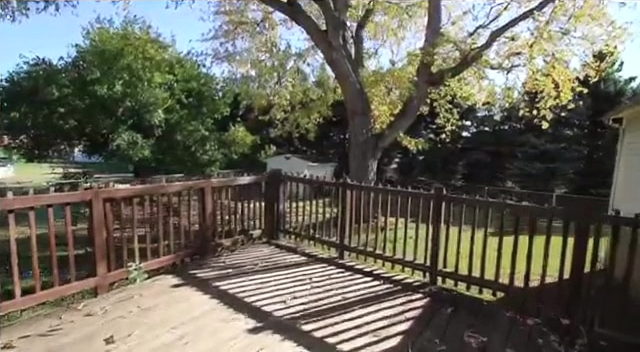} &
\includegraphics[width=0.24\textwidth]{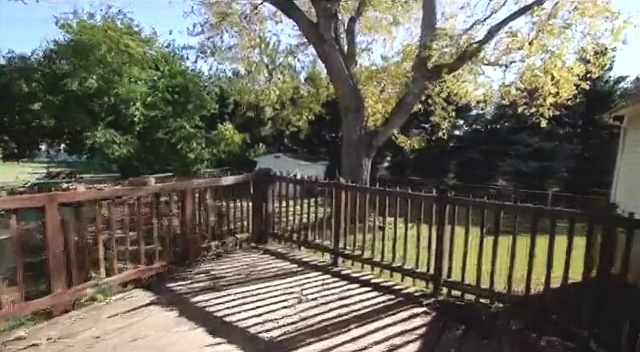} &
\includegraphics[width=0.24\textwidth]{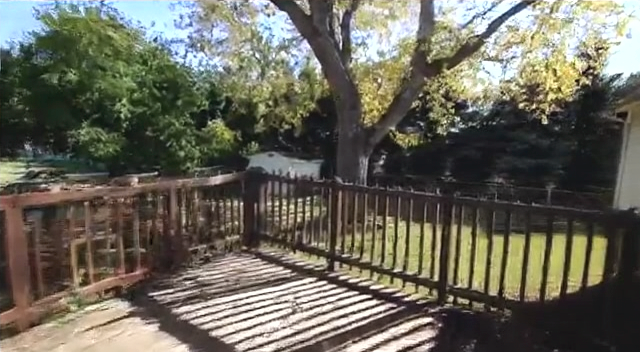} &
\includegraphics[width=0.24\textwidth]{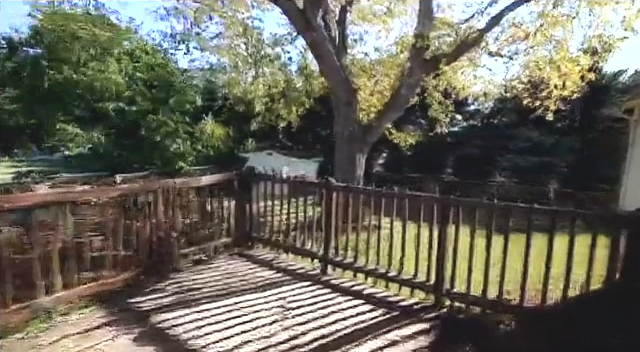} \\
View-8 & View-16 & View-24 & View-32 \\
\end{tabular}
\caption{Gac's results on outdoor scenes.}\label{fig:outdoor}
\end{figure*}

Please see Figure~\ref{fig:outdoor}.

\subsection{Interleaved depth estimation} 
\begin{figure*}[!ht]
\centering
\begin{tabular}{cccc}
\includegraphics[width=0.24\textwidth]{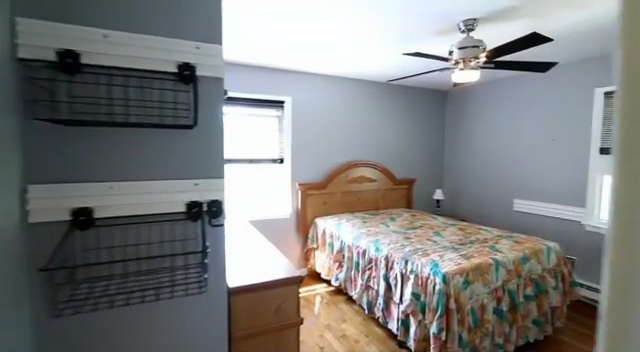} \\
Reference \\
\includegraphics[width=0.24\textwidth]{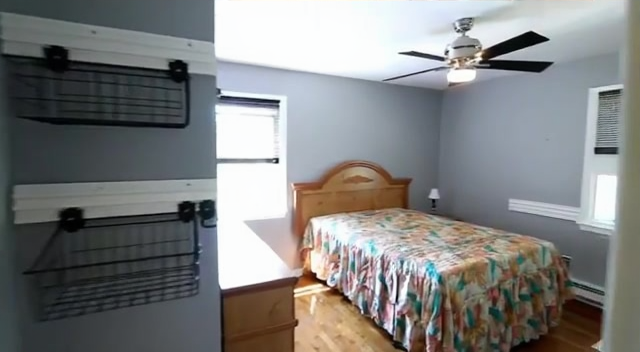} &
\includegraphics[width=0.24\textwidth]{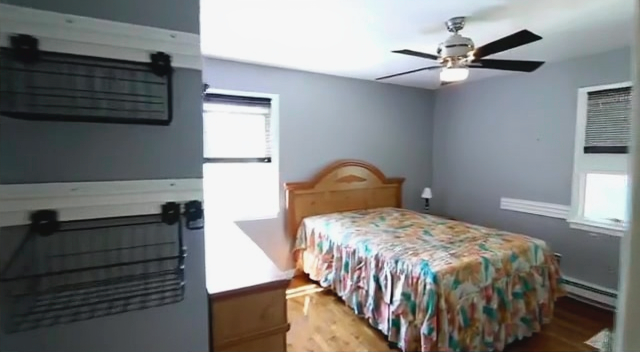} &
\includegraphics[width=0.24\textwidth]{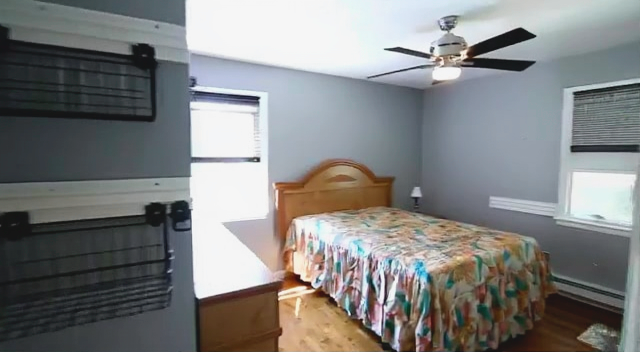} &
\includegraphics[width=0.24\textwidth]{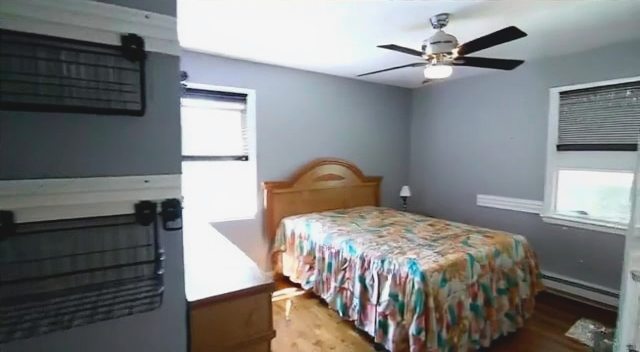} \\
View-3 & View-6 & View-9 & View-12 \\
\includegraphics[width=0.24\textwidth]{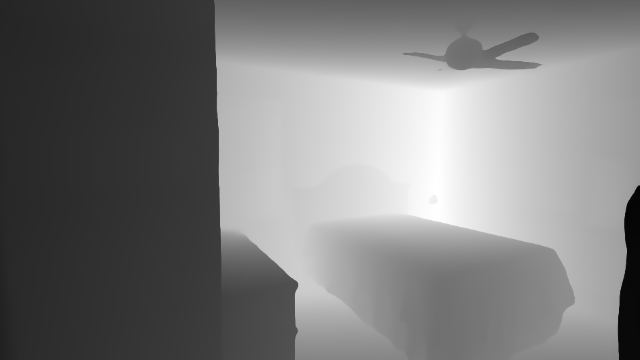} &
\includegraphics[width=0.24\textwidth]{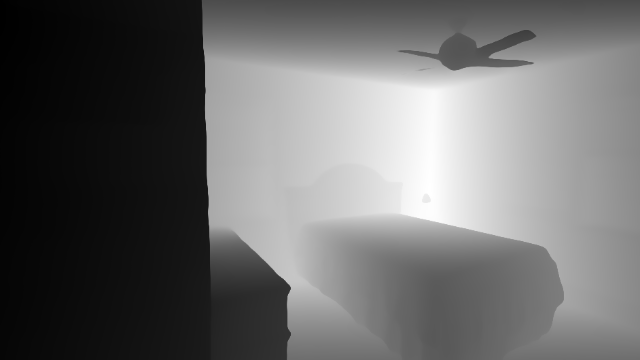} &
\includegraphics[width=0.24\textwidth]{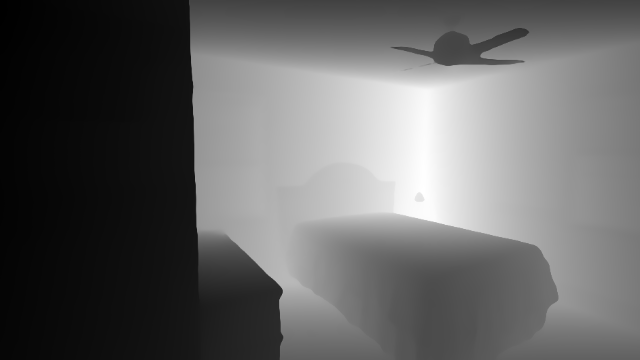} &
\includegraphics[width=0.24\textwidth]{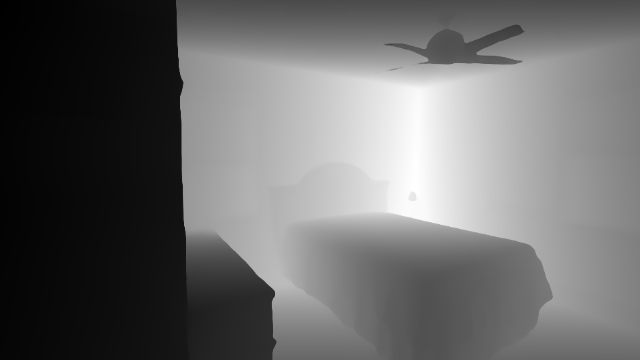} \\
Depth-3 & Depth-6 & Depth-9 & Depth-12 \\
\includegraphics[width=0.24\textwidth]{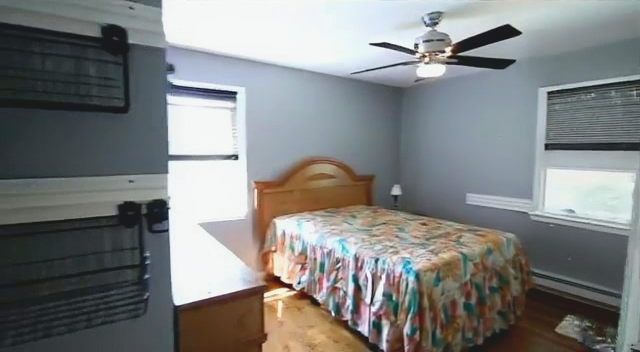} &
\includegraphics[width=0.24\textwidth]{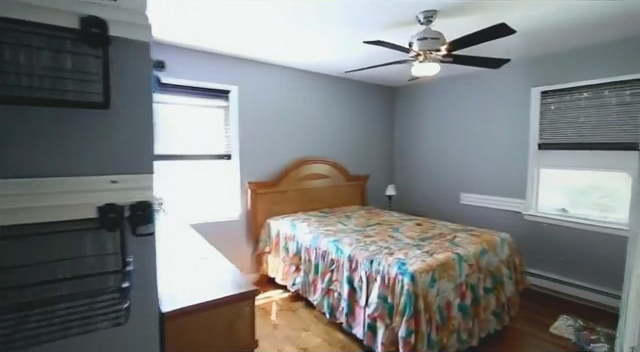} &
\includegraphics[width=0.24\textwidth]{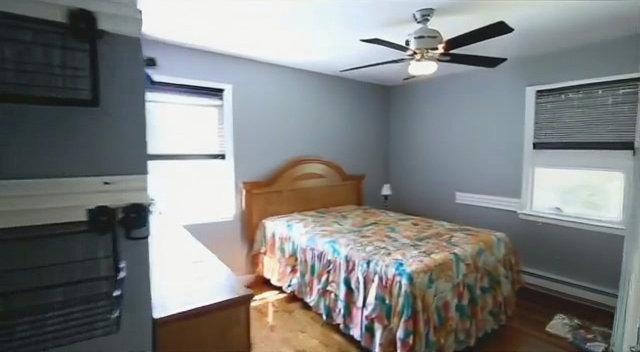} &
\includegraphics[width=0.24\textwidth]{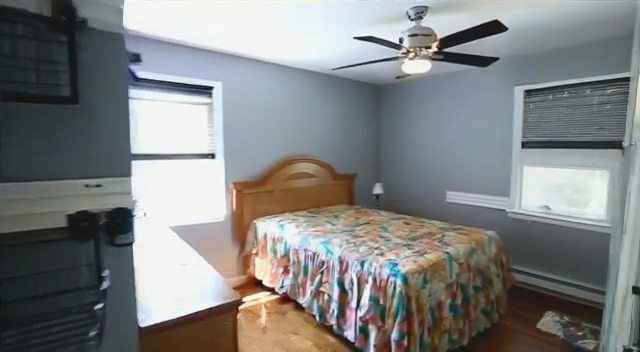} \\
View-15 & View-18 & View-21 & View-24 \\
\includegraphics[width=0.24\textwidth]{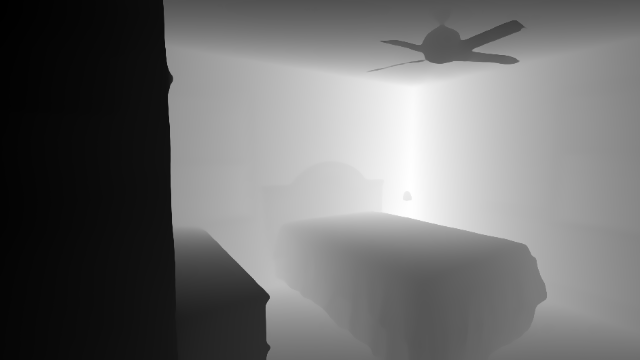} &
\includegraphics[width=0.24\textwidth]{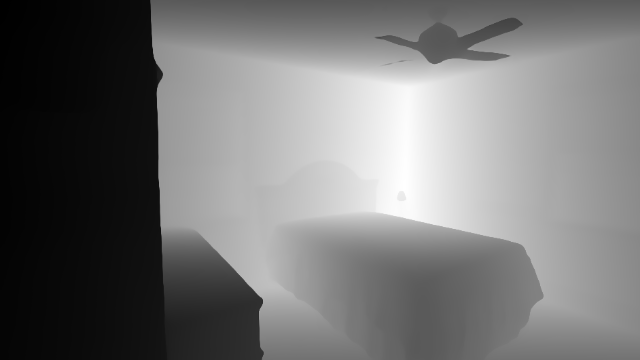} &
\includegraphics[width=0.24\textwidth]{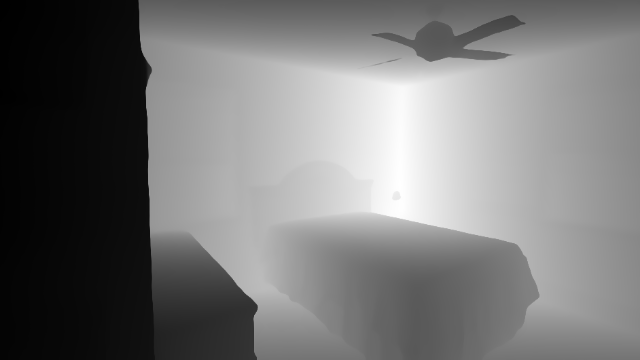} &
\includegraphics[width=0.24\textwidth]{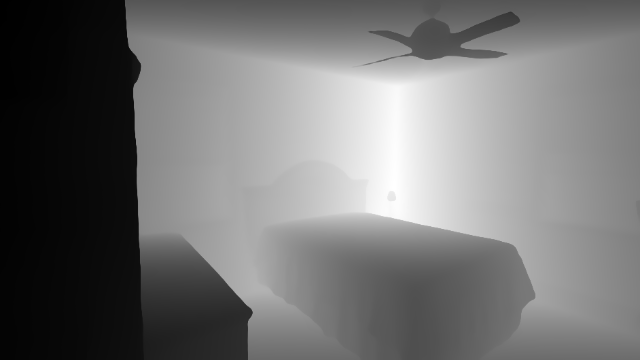} \\
Depth-15 & Depth-18 & Depth-21 & Depth-24 \\
\end{tabular}
\caption{RGB-Depth interleaved output format in indoor scenes.}\label{fig:depth_indoor}
\end{figure*}

\begin{figure*}[!ht]
\centering
\begin{tabular}{cccc}
\includegraphics[width=0.24\textwidth]{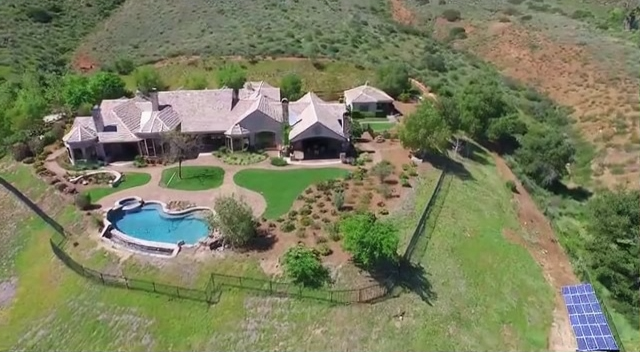} \\
Reference \\
\includegraphics[width=0.24\textwidth]{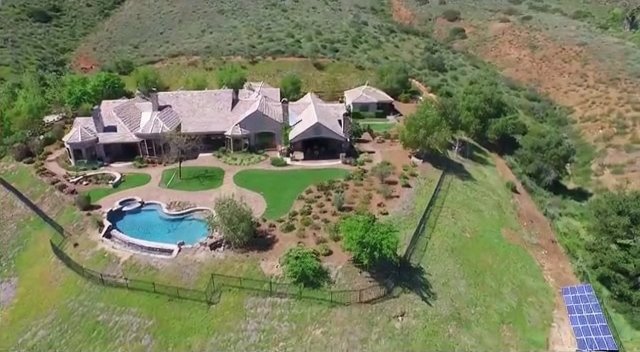} &
\includegraphics[width=0.24\textwidth]{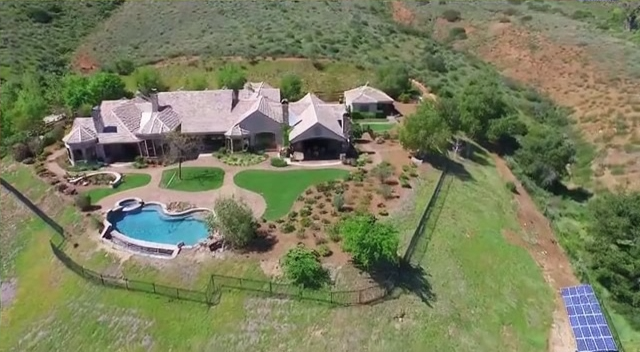} &
\includegraphics[width=0.24\textwidth]{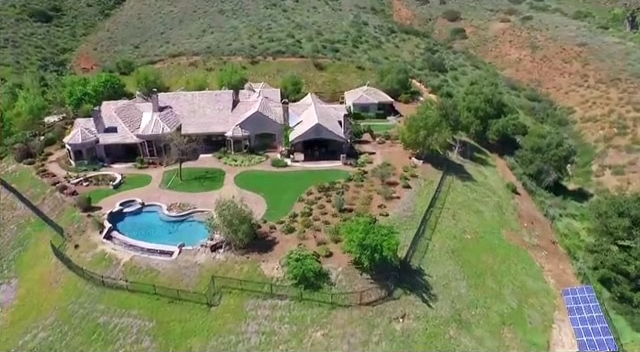} &
\includegraphics[width=0.24\textwidth]{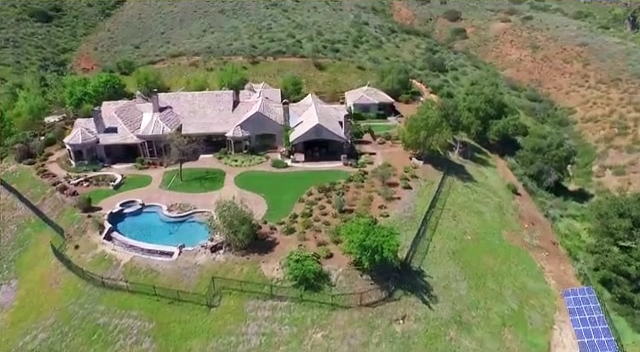} \\
View-2 & View-4 & View-6 & View-8 \\
\includegraphics[width=0.24\textwidth]{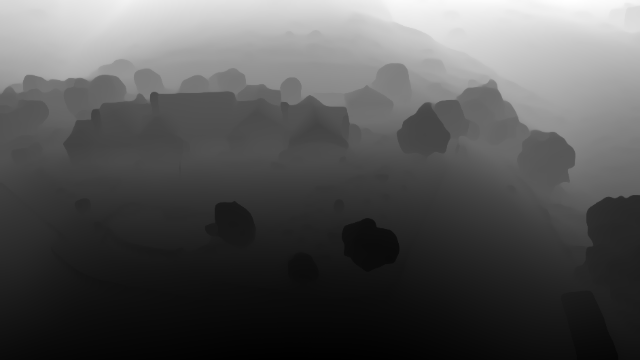} &
\includegraphics[width=0.24\textwidth]{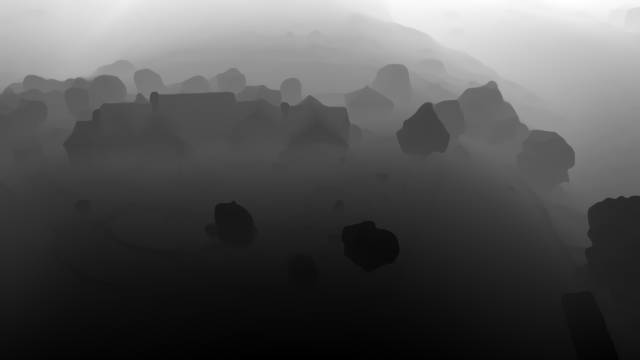} &
\includegraphics[width=0.24\textwidth]{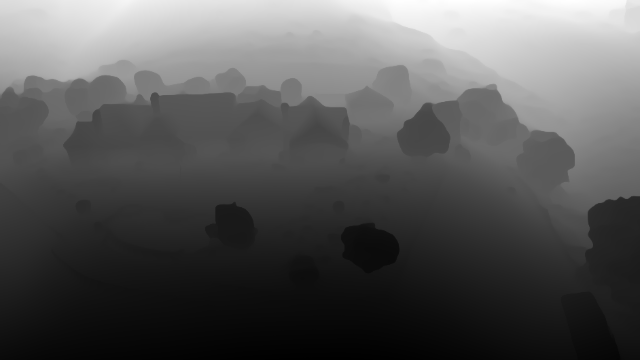} &
\includegraphics[width=0.24\textwidth]{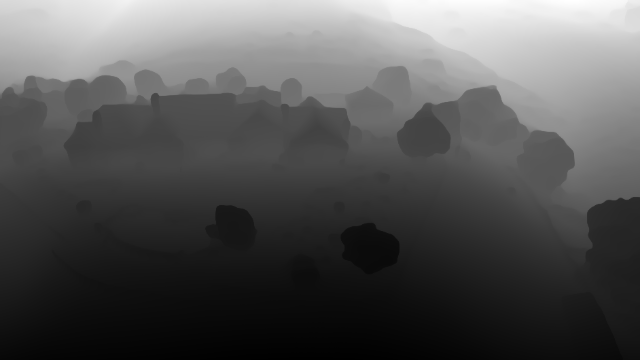} \\
Depth-2 & Depth-4 & Depth-6 & Depth-8 \\
\includegraphics[width=0.24\textwidth]{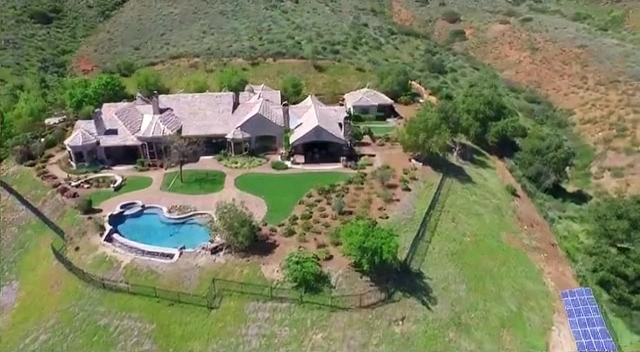} &
\includegraphics[width=0.24\textwidth]{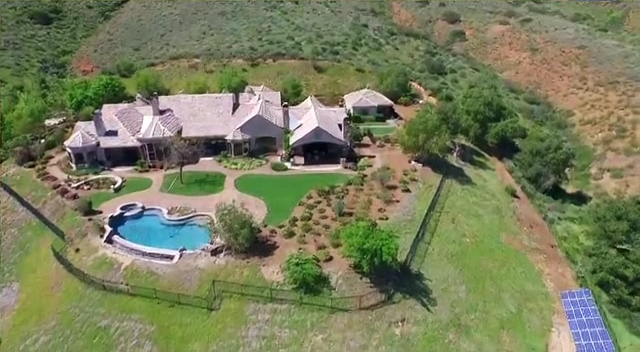} &
\includegraphics[width=0.24\textwidth]{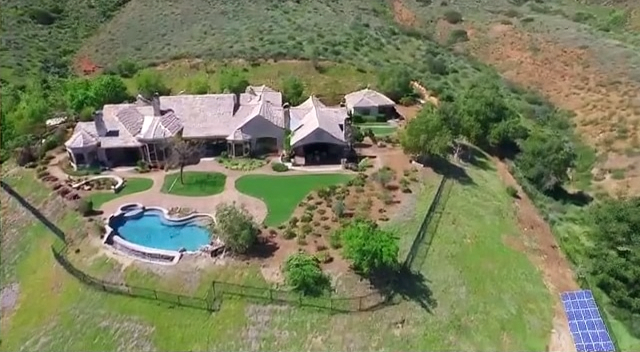} &
\includegraphics[width=0.24\textwidth]{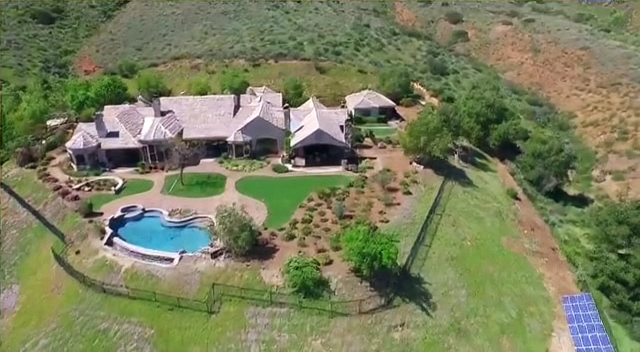} \\
View-10 & View-12 & View-14 & View-16 \\
\includegraphics[width=0.24\textwidth]{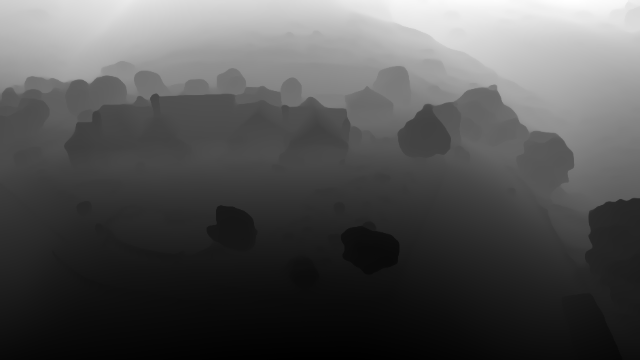} &
\includegraphics[width=0.24\textwidth]{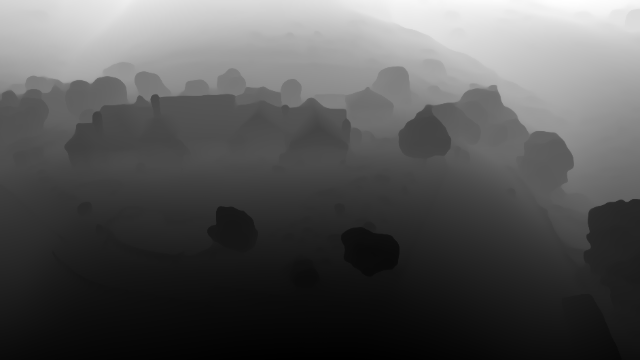} &
\includegraphics[width=0.24\textwidth]{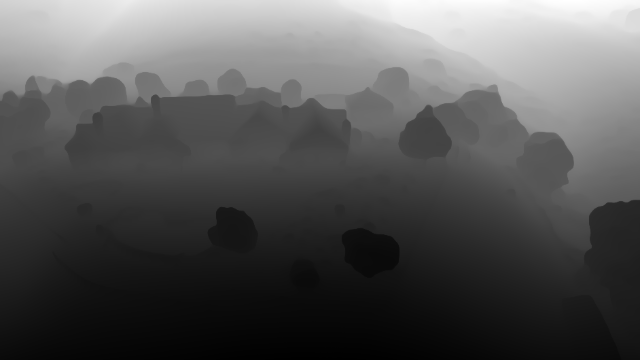} &
\includegraphics[width=0.24\textwidth]{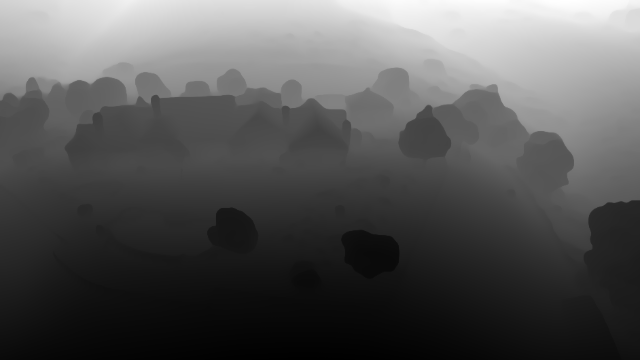} \\
Depth-10 & Depth-12 & Depth-14 & Depth-16 \\
\end{tabular}
\caption{RGB-Depth interleaved output format in outdoor scenes.}\label{fig:depth_outdoor}
\end{figure*}

Please see Figure~\ref{fig:depth_indoor} and Figure~\ref{fig:depth_outdoor}.

\subsection{Failure cases} 

\begin{figure*}[!ht]
\centering
\begin{tabular}{cccc}
\includegraphics[width=0.24\textwidth]{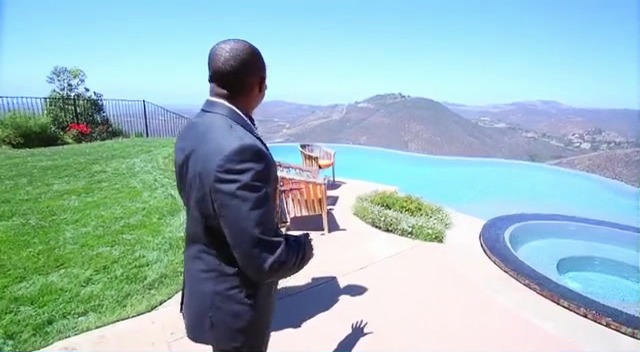} \\
Reference \\
\includegraphics[width=0.24\textwidth]{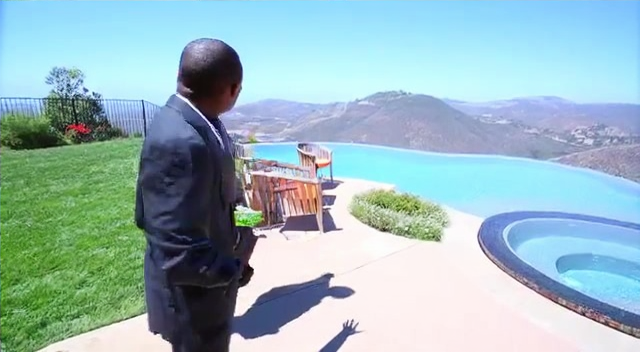} &
\includegraphics[width=0.24\textwidth]{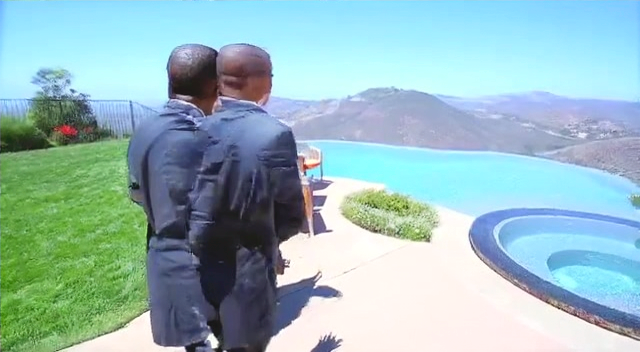} &
\includegraphics[width=0.24\textwidth]{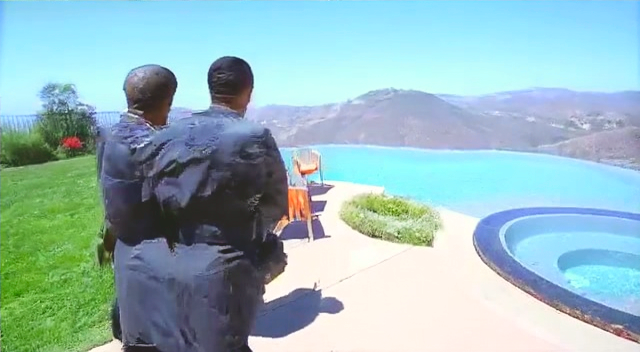} &
\includegraphics[width=0.24\textwidth]{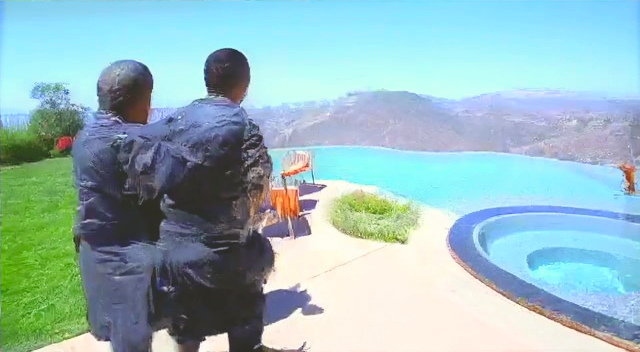} \\
View-4 & View-8 & View-12 & View-16 \\

\includegraphics[width=0.24\textwidth]{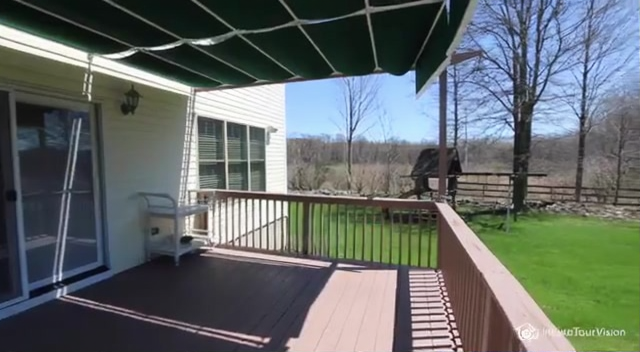} \\
Reference \\
\includegraphics[width=0.24\textwidth]{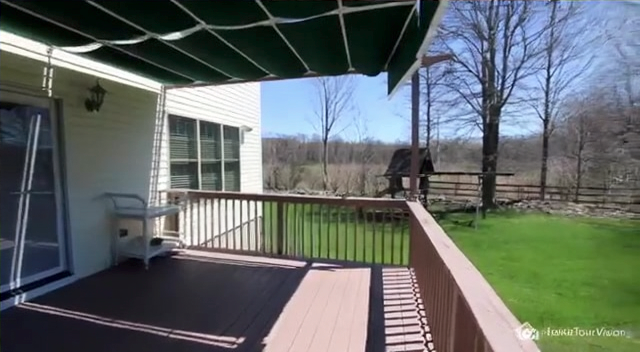} &
\includegraphics[width=0.24\textwidth]{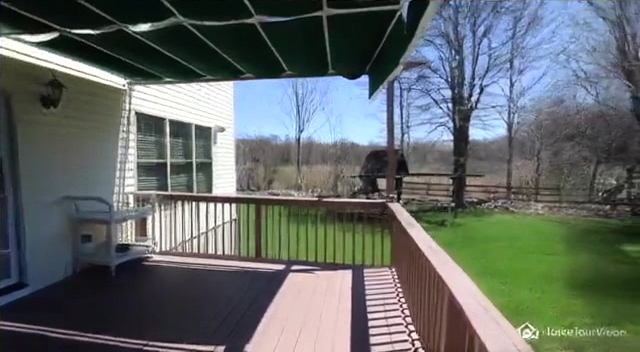} &
\includegraphics[width=0.24\textwidth]{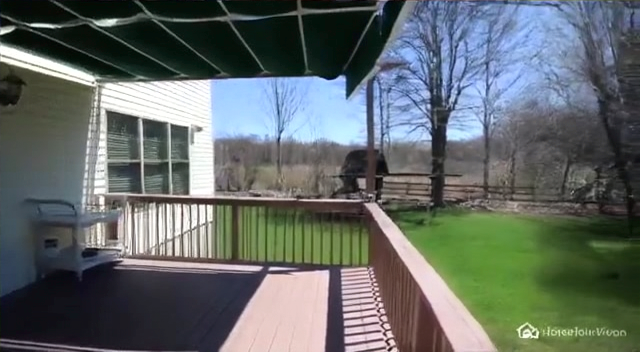} &
\includegraphics[width=0.24\textwidth]{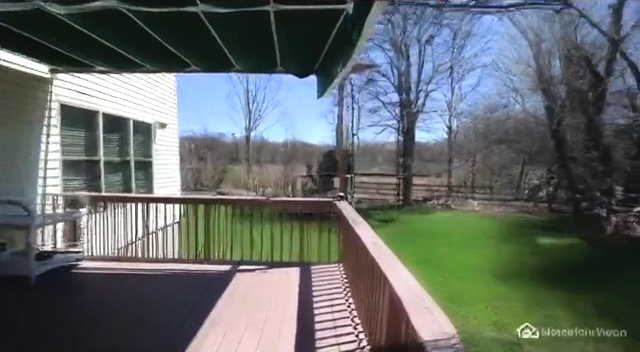} \\
View-8 & View-16 & View-24 & View-32 \\
\end{tabular}
\caption{Failure cases.}\label{fig:failure}
\end{figure*}

We present two failure cases of GaC in Figure~\ref{fig:failure}. As can be seen, GaC has not yet generalized well to humans and complex subjects. In addition, for some outdoor scenes, novel view synthesis may produce slightly darkened textures around the boundaries.

\section{Improved depth estimation strategy}

Specifically, we make two modifications to the deep prediction pipeline of Hunyuan-Voyager:

\begin{itemize}
\item We use MoGe2 for metric-scale depth prediction and incorporate the VGGT depth confidence map to refine the depth map.
\item We extend most of the operators in the pipeline from numpy to pytorch to support parallelized depth prediction on the GPU. 
\end{itemize} 

The relevant code will be open-sourced upon publication.




\end{document}